\documentclass{article}
\usepackage{graphicx} 
\usepackage{caption}
\usepackage[left]{lineno} 
\usepackage[a4paper, margin=1in]{geometry} 
\usepackage{rotating}
\usepackage [latin1]{inputenc}
\usepackage{booktabs}
\usepackage{multirow}
\usepackage{gensymb}
\usepackage[backend=bibtex,
            style=numeric,
            sorting=none,
            giveninits=true,
            maxnames=5,
            minnames=1,
            doi=false,
            isbn=false,
            url=false
]{biblatex}

\DeclareFieldFormat[article,inproceedings,online]{title}{#1}
\renewbibmacro{in:}{}

\DeclareFieldFormat{journaltitle}{\mkbibemph{#1}}
\DeclareFieldFormat{booktitle}{\mkbibemph{#1}}
\DeclareFieldFormat[article]{volume}{\textbf{#1}}

\DeclareFieldFormat{pages}{#1}
\DeclareFieldFormat{url}{#1}

\DeclareBibliographyDriver{article}{%
  \usebibmacro{bibindex}%
  \usebibmacro{begentry}%
  \usebibmacro{author/translator+others}%
  \setunit{\printdelim{nametitledelim}}\newblock
  \usebibmacro{title}%
  \newunit
  \mkbibemph{\printfield{journaltitle}}%
  \setunit{\addspace}%
  \printfield{volume}%
  \iffieldundef{pages}
    {}
    {\setunit{\addcomma\space}%
     \printfield{pages}}%
  \setunit{\space}%
  \printtext[parens]{\printfield{year}}%
  \newunit\newblock
  \usebibmacro{finentry}}

\DeclareBibliographyDriver{inproceedings}{%
  \usebibmacro{bibindex}%
  \usebibmacro{begentry}%
  \usebibmacro{author/translator+others}%
  \setunit{\printdelim{nametitledelim}}\newblock
  \usebibmacro{title}%
  \newunit\newblock
  \mkbibemph{\printfield{booktitle}}%
  \setunit{\addcomma\space}%
  \printfield{pages}%
  \setunit{\space}%
  \printtext[parens]{\printfield{year}}%
  \newunit\newblock
  \usebibmacro{finentry}}

\DeclareBibliographyDriver{online}{%
  \usebibmacro{bibindex}%
  \usebibmacro{begentry}%
  \usebibmacro{author/translator+others}%
  \setunit{\printdelim{nametitledelim}}\newblock
  \usebibmacro{title}%
  \newunit\newblock
  \iffieldundef{url}
    {\printtext[parens]{\printfield{year}}}
    {\printtext{Available at: }%
     \printfield{url}%
     \setunit{\space}%
     \printtext[parens]{\printfield{year}}}%
  \usebibmacro{finentry}}

\AtEveryBibitem{
  \clearfield{note}
  \clearfield{number}
  \clearfield{month}
  \clearfield{issue}
  \clearfield{issn}
  \clearfield{isbn}
  \clearfield{urldate}
  \clearfield{language}
  
  \iffieldundef{pages}{}{%
    \clearfield{number}%
  }
  
  \iffieldundef{url}{%
    \clearfield{urlyear}%
  }{}
}

\AtBeginBibliography{%
}

\renewbibmacro*{journal}{%
  \iffieldundef{journaltitle}
    {}
    {\printfield{journaltitle}%
     \setunit{\addcomma\space}}}

\DeclareFieldFormat[inproceedings]{booktitle}{\mkbibemph{#1}}

\usepackage{breqn}
\usepackage{breqn}
\usepackage{xcolor}
\usepackage{amssymb}     
\usepackage{amsfonts}    
\usepackage{mathtools}   
\usepackage{bm}          
\usepackage{algorithm}        
\usepackage{algorithmic}      
\usepackage{listings}         
\usepackage{booktabs}    
\usepackage{array}       
\usepackage{multirow}    
\usepackage{tabularx}    
\usepackage{textcomp}

\usepackage{graphicx} 
\usepackage[left]{lineno} 
\usepackage[a4paper, margin=1in]{geometry} 
\usepackage{amsmath}     
\usepackage{amssymb}     
\usepackage{amsfonts}    
\usepackage{mathtools}   
\usepackage{bm}          
\usepackage{algorithm}        
\usepackage{algorithmic}      
\usepackage{listings}         
\usepackage{booktabs}    
\usepackage{array}       
\usepackage{multirow}    
\usepackage{tabularx}    
\usepackage{setspace}    

\date{} 

\begin{document}

\newcommand{\suppfigure}[1]{Supplementary Figure~S#1}
\newcommand{\supptable}[1]{Supplementary Table~S#1}

\begin{flushleft}
    \Large \textbf{Spatially-Heterogeneous Causal Bayesian Networks for Seismic Multi-Hazard Estimation: A Variational Approach with Gaussian Processes and Normalizing Flows} \\
    \vspace{0.5cm}
    \normalsize
    Xuechun Li\textsuperscript{1}, Shan Gao\textsuperscript{1}, Runyu Gao\textsuperscript{1}, Susu Xu\textsuperscript{1,2,*} \\
    \textsuperscript{1}\textit{Center for Systems Science and Engineering, Department of Civil and Systems Engineering, Johns Hopkins University}\\
    \textsuperscript{2}\textit{Data Science and AI Institute, Johns Hopkins University} \\
    \textsuperscript{*}\textit{Corresponding author: Susu Xu, email: \url{susuxu@jhu.edu}}
\end{flushleft}

\noindent Post-earthquake hazard and impact estimation are critical for effective disaster response, yet current approaches face significant limitations. Traditional models employ fixed parameters regardless of geographical context, misrepresenting how seismic effects vary across diverse landscapes, while remote sensing technologies struggle to distinguish between co-located hazards. We address these challenges with a spatially-aware causal Bayesian network that decouples co-located hazards by modeling their causal relationships with location-specific parameters. Our framework integrates sensing observations, latent variables, and spatial heterogeneity through a novel combination of Gaussian Processes with normalizing flows, enabling us to capture how same earthquake produces different effects across varied geological and topographical features. Evaluations across three earthquakes demonstrate Spatial-VCBN achieves Area Under the Curve (AUC) improvements of up to 35.2\% over existing methods. These results highlight the critical importance of modeling spatial heterogeneity in causal mechanisms for accurate disaster assessment, with direct implications for improving emergency response resource allocation.

\section{Introduction}\label{sec1}

Earthquakes cause harm not only through direct ground shaking but also by triggering secondary ground failures such as landslides and liquefaction. These combined effects lead to devastating consequences, including structural damage and human casualties. A striking illustration is the 2021 Haiti earthquake, which initiated over 7,000 landslides covering more than 80 square kilometers. This catastrophic event resulted in damage or destruction to over 130,000 buildings, claimed 2,248 lives, and left more than 12,200 people injured \cite{havenith2022first}. Rapidly identifying where and how severely ground failures and structural damage have occurred following an earthquake is essential for effective victim rescue operations within the crucial "Golden 72 Hour" window, and plays a vital role in developing effective post-disaster recovery plans \cite{jang2009rescue, li2023disasternet}.

Over the years, researchers have developed various approaches for estimating the location and intensity of earthquake-induced ground failures and building damage. Traditional approaches to seismic hazard assessment fall into two main categories: physical and statistical models \cite{toprak2003liquefaction,marc2016seismologically, newmark1965effects,ghofrani2014site}. Physical models apply fundamental geotechnical principles, such as the Newmark displacement method for landslides or liquefaction potential indices. While foundational, these approaches require detailed geotechnical data often unavailable during rapid response and frequently oversimplify complex physical dynamics. Statistical models estimate hazards using geospatial susceptibility indicators paired with ground motion data, calibrated on historical events. However, current approaches employ fixed parameters regardless of location. This one-size-fits-all approach misrepresents how seismic effects vary across different geological compositions, topographical features, and infrastructure, leading to significant prediction inaccuracies. 

Remote sensing technologies, particularly Interferometric Synthetic Aperture Radar (InSAR), have revolutionized rapid post-earthquake assessment capabilities. InSAR works by comparing phase differences between radar signals captured at different times, enabling detection of surface deformation with high-resolution precision \cite{Yun2015Rapid}. Damage Proxy Maps (DPMs) generated by the NASA's Advanced Rapid Imaging and Analysis (AIRA) team, which is a state-of-the-art InSAR product, identify ground changes by analyzing radar measurement disparities before and after seismic events \cite{fielding2024damage}. Although remote sensing technologies facilitate expedited hazard evaluation \cite{yu2024intelligent}, these approaches typically address individual hazards and face challenges in differentiating between overlapping signals from multiple hazard types while filtering out environmental noise \cite{kongar2017evaluating}. This limitation creates a critical information gap for emergency responders who need a comprehensive understanding of multiple concurrent hazards.

Recent advances have introduced causal Bayesian networks to address the limitations of remote sensing technologies by disentangling cascading earthquake impacts \cite{xu2022seismic, xucausality, li2025rapid, wang2024scalable, wang2023causality, li2023m7, li2023normalizing, li2024optimizing}. These probabilistic models leverage variational inference to track causal chains from initial earthquake events through multiple hazards to ultimate building damage. A recent approach \cite{li2024spatial} attempted to account for spatial variations using bilateral filters, which is a technique that creates weighted averages based on proximity between locations \cite{Tomasi1998BilateralFF, Paris2009BilateralFT}. While this improved upon uniform-parameter models, bilateral filtering fundamentally treats spatial dependency as a simple averaging operation without modeling how the underlying causal mechanisms themselves vary across different geological settings \cite{Gelfand2016SpatialSA, Diggle2013ModelbasedG}. For example, the same level of ground shaking might have a much stronger causal effect on triggering landslides in areas with steep slopes and loose soil compared to flat areas with stable bedrock \cite{Xu2012ComparisonOD}. Similarly, building damage resulting from liquefaction will vary dramatically depending on foundation types and subsurface conditions. These location-specific causal relationships cannot be captured by simple spatial averaging \cite{Banerjee2003HierarchicalMA}.

To address these limitations, we propose a spatially-aware variational causal Bayesian network (Spatial-VCBN) that represents a fundamental shift in approach: rather than modeling spatial correlation as an afterthought, we directly model the variation in causal mechanisms themselves. The distinction is crucial because while previous approaches might apply spatial smoothing to already-estimated parameters, our framework incorporates spatial heterogeneity directly into the causal structure itself. Such integration captures the reality that the same earthquake generates dramatically different effects depending on local conditions. The physical intuition behind Spatial-VCBN is that causal relationships in earthquake scenarios are fundamentally location-dependent, a principle well-established in geophysics but rarely incorporated into predictive models. While simpler approaches like linear spatial interpolation or kernel smoothing might seem sufficient, the highly non-linear and potentially multi-modal nature of earthquake causal effects demands more sophisticated methods. Our framework consists of three key components: (1) observable variables including geospatial features and damage proxy maps derived from satellite imagery; (2) latent hazard/impact variables representing unobserved landslides, liquefaction, and building damage; and (3) spatially-varying causal coefficients that quantify the strength of causal relationships at each location. 

A key innovation in Spatial-VCBN is the modeling of these spatially-varying causal coefficients using a combination of Gaussian Processes (GPs) with normalizing flows. GPs serve as spatial priors that ensure causal relationships vary smoothly across regions based on geophysical similarities \cite{ray2019bayesian}, while normalizing flows provide flexible, non-linear transformations that capture complex \cite{rezende2015variational, li2023disasternet, li2023normalizing}, non-Gaussian distributions of causal effects. Conceptually, normalizing flows transform simple probability distributions into more complex ones through a sequence of invertible mappings, allowing us to represent the richly varied ways that earthquake forces translate into surface effects across different terrains. The use of Gaussian Processes is particularly well-suited for this problem because they naturally model spatial correlation while allowing for location-specific variations. Normalizing flows complement this approach by transforming simple distributions into more complex ones through a series of invertible mappings, enabling Spatial-VCBN to capture the highly non-linear and potentially multi-modal nature of causal effects in earthquake scenarios. Our ablation studies reveal that an optimal flow number of $K=6$ is required to adequately model these complex relationships, confirming that simpler distributional assumptions would fail to capture the physical reality of earthquake impacts. This is crucial because the relationships between geological features, seismic waves, and resulting hazards often follow complex, non-Gaussian patterns that simpler distributional assumptions would fail to capture. 

By combining GPs with normalizing flows, our framework can represent complex, non-Gaussian distributions of causal effects that better reflect the physical reality of how earthquake impacts propagate through diverse environments. This probabilistic approach also enables principled quantification of uncertainty in hazard estimates, which is crucial for effective disaster response planning. For inference, we implement a stochastic variational approach with an expectation-maximization algorithm that alternates between updating posteriors of unobserved variables and refining model parameters. To handle large geographical regions efficiently, we apply a local pruning strategy that exploits the natural sparsity in real-world causal networks, achieving processing speeds of approximately 0.94 seconds per $km^{2}$ with GPU acceleration, making Spatial-VCBN viable even for large-scale events like the Haiti earthquake (15,970 $km^{2}$).

Our empirical evaluations demonstrate that this spatially-variant approach significantly improves hazard and impact estimation accuracy, with AUC improvements of up to 35.2\% over prior probability baselines and 5.5\% over state-of-the-art methods across three earthquake events (Haiti, Puerto Rico, and Turkey-Syria). These substantial improvements are not merely incremental advances but represent a step-change in our ability to rapidly assess complex disaster scenarios with the precision needed for effective emergency response. The model shows remarkable robustness in signal-constrained environments, successfully extracting coherent hazard patterns even from noisy DPM signals, as demonstrated in mountainous and coastal regions. Particularly noteworthy is the strong performance of Spatial-VCBN at low false positive rates, which is crucial for effective resource allocation in disaster response operations where false alarms can waste limited resources. These results highlight the critical importance of modeling spatial heterogeneity in the actual causal mechanisms themselves, rather than merely accounting for spatial proximity, for effective disaster assessment and response planning.

\section{Methods}\label{sec4}

\subsection{Prior Geospatial Models}

Earthquake shaking intensity serves as the fundamental catalyst for structural damage and geohazards like landslides and liquefaction. We incorporate this critical factor through USGS ShakeMap data, which provides comprehensive shaking metrics. The composite ShakeMap effectively captures maximum ground motion intensities, enabling more accurate assessments of structural and geologic impacts, which is an invaluable resource for post-earthquake forensic analysis.

\begin{figure}[htbp]
\begin{center}
\scalebox{1}{
\begin{tabular}{c}
\includegraphics[width=1\columnwidth]{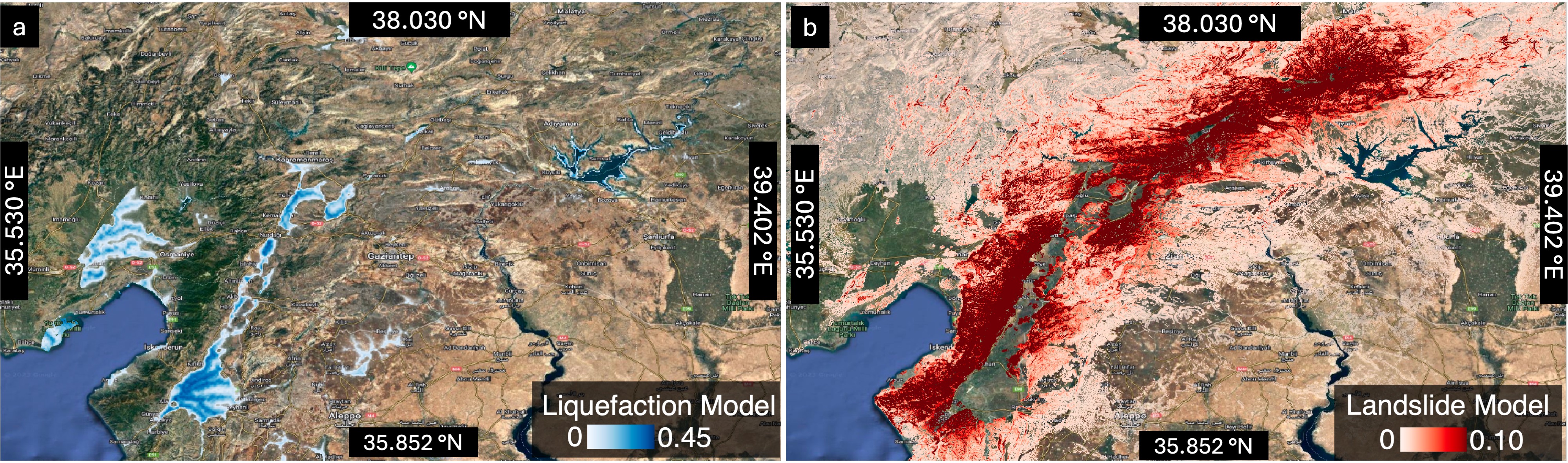}
\end{tabular}}
\caption{After the 2023 Turkey-Syria earthquake sequence, the USGS produced example ground failure models for landslide and liquefaction \cite{Fig3}. The probability of ground failure models is what the legend colors represent.}
\label{Turkey_PLSLF}
\vspace{-0.5cm}
\end{center}
\end{figure}

Beyond basic shaking information, USGS expanded their rapid post-earthquake information products in 2018 to include ground-failure estimations \cite{wald2003shakemap}. These specialized models predict the likelihood and distribution of earthquake-triggered landslides and liquefaction \cite{allstadt2022ground}. The modeling framework integrates multiple environmental factors including terrain slope, geological susceptibility, and soil characteristics to generate probability maps \cite{zhu2016updated,nowicki2018global}. An illustrative example appears in Figure \ref{Turkey_PLSLF}, showing the USGS ground failure projections following the 2023 Turkey-Syria earthquake. Within our Bayesian network architecture (Figure \ref{BN}), these established USGS ground failure products are incorporated as prior probability distributions for landslide and liquefaction occurrence, represented by nodes $\alpha_{LS}$ and $\alpha_{LF}$ respectively.

\subsection{InSAR Data and Damage Proxy Maps}

\begin{figure}[htbp]
\begin{center}
\scalebox{1}{
\begin{tabular}{c}
\includegraphics[width=\columnwidth]{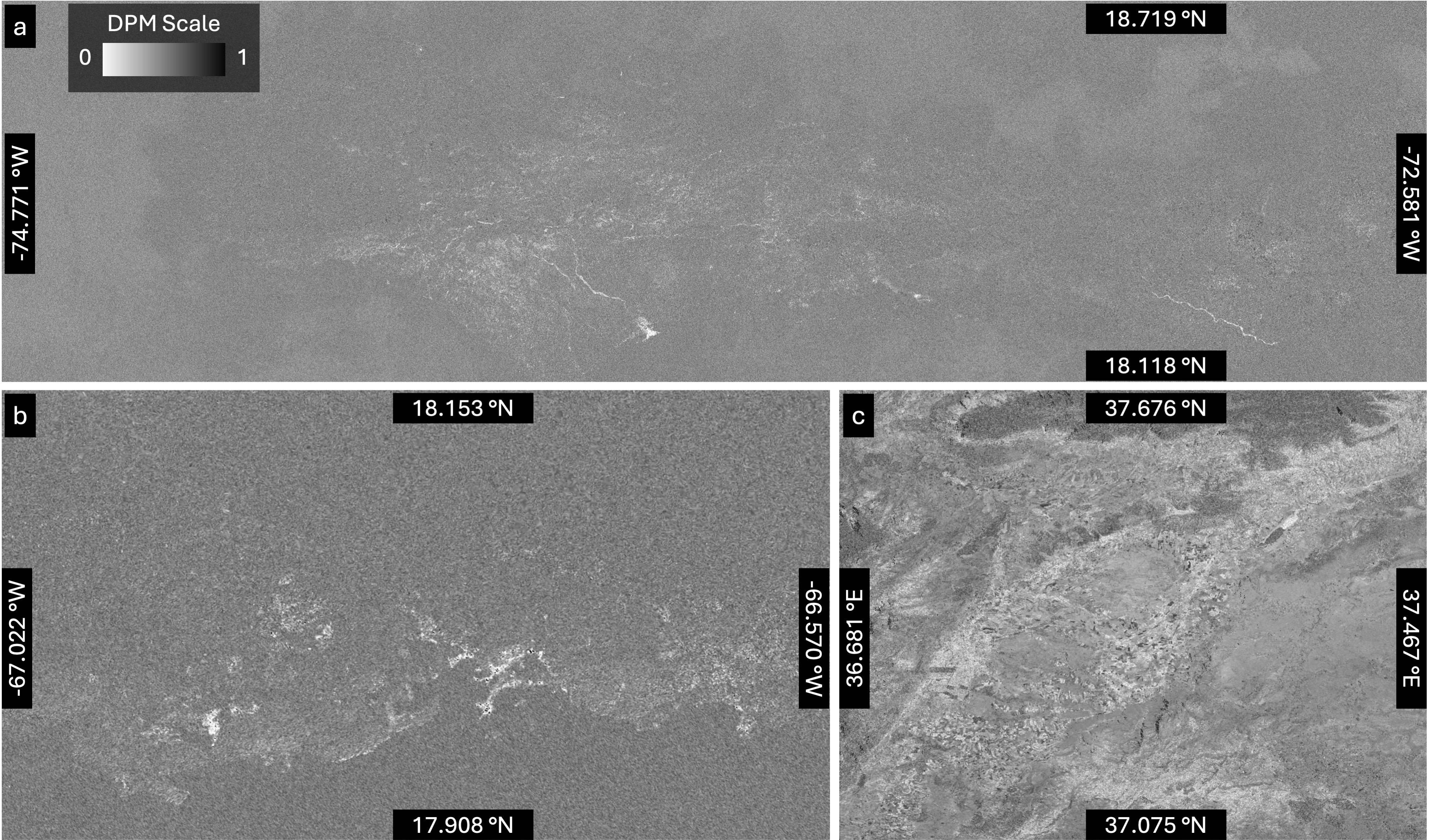}
\end{tabular}}
\caption{Damage proxy maps (DPM) generated by the AIRA team after (a) Haiti, (b) Puerto Rico, (c) Turkey-Syria earthquakes.  DPMs show surface change detection with brighter areas indicating greater surface deformation. The areas covered by DPMs are also our study regions.}
\label{DPMs}
\vspace{-0.5cm}
\end{center}
\end{figure}

Following the 2020 Puerto Rico earthquake, the 2021 Haiti earthquake, and the 2023 Turkey-Syria earthquake sequence, the ARIA team \cite{ARIA1} generated a DPM utilizing SAR images taken by the Sentinel-1A satellite operated by the European Space Agency (ESA). The DPM we employ is shown in Figure \ref{DPMs}. We utilize DPM as our sensing observations because they provide high-resolution (typically 30-meter pixels or finer) pre- and post-earthquake surfacial changes related to earthquake-induced ground failure and building damage, despite their inherent noise characteristics. The coverage of the DPMs is (a) from $-74.771 \degree W$ to $-72.581 \degree W$, from $18.118 \degree N$ to $18.719 \degree N$ for the 2021 Haiti earthquake (Figure \ref{DPMs}(a)); (b) from $-67.022 \degree W$ to $-66.570 \degree W$, from $17.908 \degree N$ to $18.153 \degree N$ for the 2020 Puerto Rico earthquake (Figure \ref{DPMs}(b)); (c) from $36.681 \degree E$ to $37.467 \degree E$, from $37.075 \degree N$ to $37.676 \degree N$ for the 2023 Turkey-Syria earthquake sequence (Figure \ref{DPMs}(a)).

\subsection{Geophysical Features}
We incorporate six geophysical features in Spatial-VCBN: (1)~\emph{Vs30}, (2)~\emph{Slope from DEM}, (3)~\emph{Land Cover}, (4)~\emph{DEM}, (5)~\emph{CTI (Compound Topographic Index)}, and (6)~\emph{Coast Distance}. Below, we provide a brief rationale for including each feature.

\begin{itemize}
    \item \textbf{Vs30 (Shear-Wave Velocity over 30\,m):} Vs30 is widely recognized as a key parameter for characterizing local site conditions and capturing near-surface amplification effects in seismic hazard assessments. Higher Vs30 values generally correspond to stiffer soils or rocks, reducing the likelihood of significant ground amplification \cite{wald1999relationships, seyhan2014semi}.

    \item \textbf{Slope from DEM:} Slope, derived from a Digital Elevation Model (DEM), is crucial in identifying areas susceptible to landslides and other mass-movement hazards triggered by seismic activity. Regions with steep slopes tend to experience higher landslide hazard potential \cite{kamp2008gis, danielson2011global}.

    \item \textbf{Land Cover:} Land cover information helps discern the presence of vegetation, urban infrastructure, or water bodies. Different land cover classes (e.g., dense vegetation vs.\ built-up areas) can significantly influence ground response and post-earthquake soil stability \cite{arino2012global}.

    \item \textbf{DEM (Elevation):} The base elevation data from DEM are essential for capturing broader topographic variations. Elevation can affect not only slope angle but also local climate, drainage patterns, and material properties, all of which have implications for earthquake-induced geophysical hazards \cite{farr2007shuttle}.

    \item \textbf{CTI (Compound Topographic Index):} CTI, sometimes referred to as the Topographic Wetness Index, quantifies the potential for water accumulation in the terrain. High CTI values often indicate persistently wetter soils, influencing both liquefaction susceptibility and potential post-seismic runoff or slope failure \cite{beven1979physically}.

    \item \textbf{Water Body Distance:} Proximity to the Water Body plays a significant role in liquefaction evaluations because saturated coastal regions and near-shore sediments are more prone to ground failure under strong seismic shaking \cite{lehner2006hydrosheds, zhu2016updated}.

    \item \textbf{Lithology:} Lithology plays a crucial role in assessing earthquake-related hazards as it directly influences seismic wave propagation, amplification, and ground shaking intensity. Different rock types exhibit varying mechanical properties such as stiffness, strength, and porosity, which affect the degree of energy dissipation and resonance during an earthquake. Additionally, lithological features control the susceptibility of slopes to landslides and liquefaction in earthquake-prone areas. Understanding lithology is therefore essential for accurate hazard zoning and designing resilient infrastructure in seismically active regions.\cite{nowicki2018global, hartmann2012new}
    
\end{itemize}
\noindent These geophysical features collectively represent critical factors governing seismic hazard distributions, spanning soil stiffness, topography, hydrology, and land use. By explicitly modeling their spatial variations and interactions, we aim to more accurately capture the localized nature of seismic impacts across heterogeneous terrains.

\begin{figure}[htbp]
\begin{center}
\scalebox{1}{
\begin{tabular}{c}
\includegraphics[width=0.6\columnwidth]{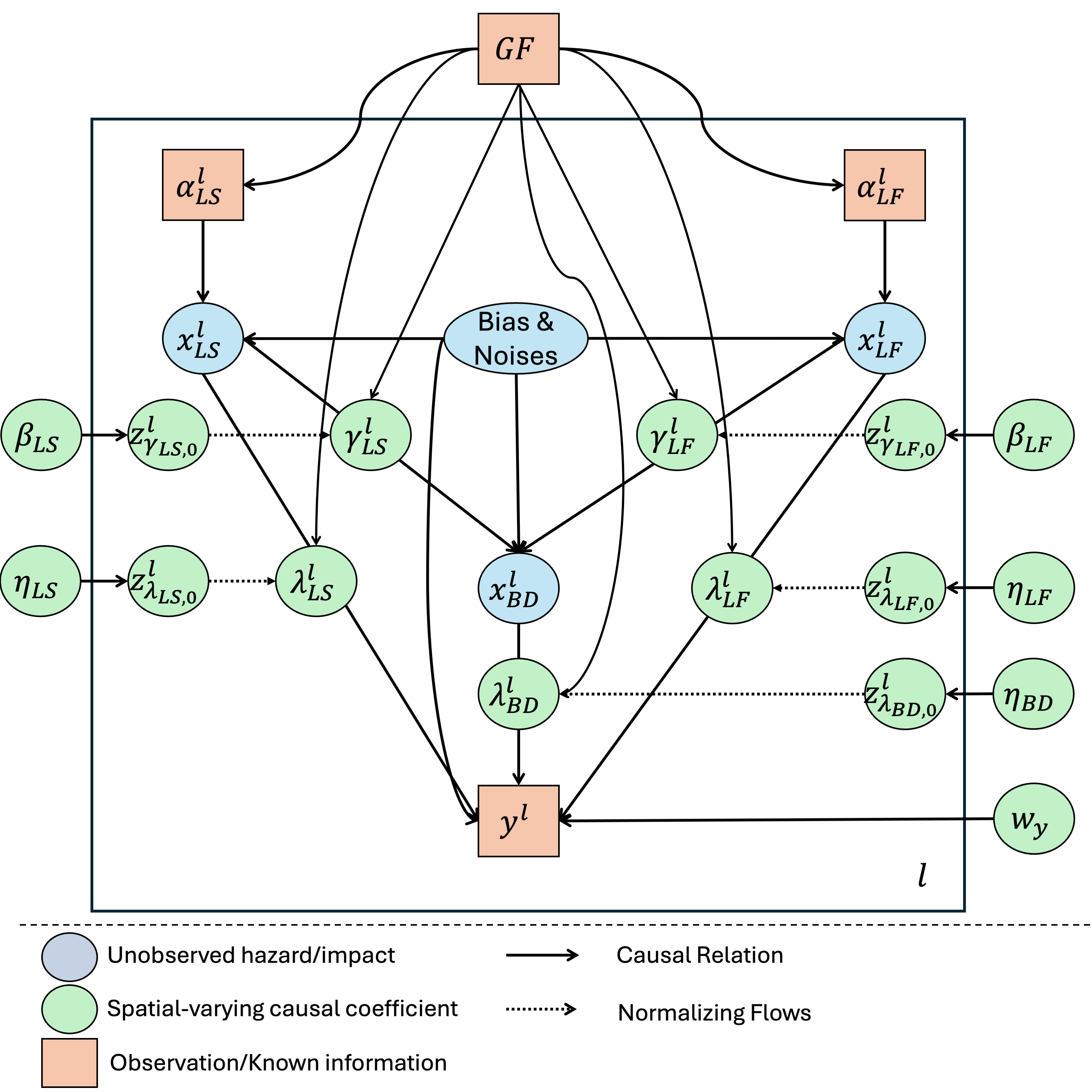}
\end{tabular}}
\caption{Overview of our causal Bayesian inference framework for seismic multi-hazard and impacts estimation. $l$ in the figure refers to the $l^{\text{th}}$ location in a target area. Blue circles refer to latent hazard/impact variables. Green cirles refer to the spatial-varying causal coefficients. Orange rectangles refer to the observations or known information. }
\label{BN}
\vspace{-0.5cm}
\end{center}
\end{figure}

\subsection{Causal Modeling for Disaster Impacts}

Disasters propagate through complex chains of causation, where initial events trigger cascading hazards that ultimately result in observable impacts. To accurately model these processes, we develop a spatially-aware causal Bayesian network that captures both the structural relationships between variables and the spatial heterogeneity of these relationships. Figure \ref{BN} presents our causal Bayesian network architecture, which consists of three key component types: (1) observable variables (represented as orange rectangles) including geospatial features (GF) and damage proxy maps (DPM); (2) latent hazard/impact variables (shown as blue circles) representing unobserved landslides (LS), liquefaction (LF), and building damage (BD); and (3) spatially-varying causal coefficients that quantify the strength of causal relationships at each location.

For each location $l$ in our study area, we define the leaf node $y^l$ as the damage proxy map observation. Its relationship with parent nodes follows a log-linear model:

\begin{equation}
\log y^{l}|\mathcal{P}(y^{l}), \epsilon_{y}^{l}, \boldsymbol{\lambda_{\mathcal{P}(y^{l})}}^{l} \sim N(w_{0y} + \sum_{k \in \mathcal{P}(y^{l})}\lambda_{k}^{l}x_{k}^{l}, w_{\epsilon_{y}}^{2})
\label{logy_assumption}
\end{equation}

\noindent where $x_k^l$ represents the hidden hazard/damage variables, $\boldsymbol{\lambda_{\mathcal{P}(y^{l})}}^{l}$ denotes the spatially-varying causal coefficients that quantify the causal relationship from latent variables $x_{k} \in \{LS, LF, BD\}$ to $y$, and $\epsilon_y$ accounts for environmental noise. The hidden nodes $x_i^l \in \{0, 1\}$ for $i \in \{LS, LF, BD\}$ are binary variables indicating hazard occurrence, with activation probabilities governed by their respective parent nodes and spatially-varying coefficients $\gamma_i^l$ that quantify the causal relationship from the parent nodes $\mathcal{P}(x)$ to $x$:
\begin{equation}
\begin{aligned}
& \log\frac{p(x_{i}^{l} = 1|x_{\mathcal{P}(x_{i}^{l})}, \epsilon_{i}^{l}, \gamma_{\mathcal{P}(x_{i}^{l})})}{1-p(x_{i}^{l} = 1|x_{\mathcal{P}(x_{i}^{l})}, \epsilon_{i}^{l}, \gamma_{\mathcal{P}(x_{i}^{l})})}  = \sum_{k \in \mathcal{P}(x_{i}^{l})} \gamma_{k}^{l}x_{k}^{l} + w_{\epsilon_{i}}\epsilon_{i} + w_{0i}\\
& p(x_{i}^{l}|x_{\mathcal{P}(x_{i}^{l})}, \epsilon_{i}^{l}, \gamma_{\mathcal{P}(x_{i}^{l})})  = [p(x_{i}^{l} = 1|x_{\mathcal{P}(x_{i}^{l})}, \epsilon_{i}^{l}, \gamma_{\mathcal{P}(x_{i}^{l})})]^{x_{i}^{l}}\times [1 - p(x_{i}^{l} = 1|x_{\mathcal{P}(x_{i}^{l})}, \epsilon_{i}^{l}, \gamma_{\mathcal{P}(x_{i}^{l})})]^{1 -x_{i}^{l}}
\end{aligned}
\end{equation}

\subsection{Modeling Spatially Heterogeneous Causal Effects Using Normalizing Flows with Gaussian Process}

A key innovation in Spatial-VCBN is the modeling of spatially-varying causal coefficients $\lambda^l$ and $\gamma^l$ using Normalizing Flows with Gaussian Process. This combination allows us to capture both the spatial correlation structure inherent in these coefficients while accommodating potentially complex, non-Gaussian posterior distributions. The following sections detail our methodology for posterior inference in this model and demonstrate its effectiveness for multi-hazard impact estimation.

To capture the complex spatial variation in causal coefficients $v^{l} \in \{\lambda_{a}^{l},\gamma_{b}^{l}\}$ where $a\in \{LS,LF, BD\}$ and $b \in \{\alpha_{LS}, \alpha_{LF}, LS, LF\}$, we propose a novel approach that combines Gaussian Processes (GPs) with normalizing flows. This approach effectively captures both the spatial correlation structure and the complex, potentially non-Gaussian distributions of causal effects.

We introduce latent spatial variables $z_{v}^{l}$ which serve as the building blocks for our spatially varying causal coefficients. These latent variables represent underlying spatial patterns that, after transformation, will yield the causal parameters used in Spatial-VCBN. By working with these latent variables rather than directly modeling the coefficients, we gain mathematical tractability while maintaining expressiveness.






We model these latent variables $z_{v}^{l}$ across different locations using a Gaussian Process prior, which naturally captures correlations between locations with similar geophysical characteristics:
\begin{equation}
\mathbf{z}_{v} \sim \mathcal{GP}(m_v(\mathbf{GF}), k_v(\mathbf{GF}, \mathbf{GF}'))
\label{GPPrior}
\end{equation}
\noindent where $\mathbf{z}_{v} = [z_{v}^{1}, z_{v}^{2}, ..., z_{v}^{L}]$ is the vector of latent variables across all locations, $m_v(\mathbf{GF})$ is the mean function that depends on the geophysical features $\mathbf{GF}$ at all locations, and $k_v(\mathbf{GF}, \mathbf{GF}')$ is the kernel function that defines the covariance structure between different geophysical features. We use a Matern kernel with smoothness parameter $\nu = 3/2$:
\begin{equation}
k_v(\mathbf{GF}_i, \mathbf{GF}_j) = \sigma^2_v \frac{2^{1-\nu}}{\Gamma(\nu)} \left(\sqrt{2\nu}\frac{||\mathbf{GF}_i - \mathbf{GF}_j||}{\ell_v}\right)^{\nu} K_{\nu}\left(\sqrt{2\nu}\frac{||\mathbf{GF}_i - \mathbf{GF}_j||}{\ell_v}\right)
\end{equation}
\noindent where $\sigma^2_v$ controls the variance, $\ell_v$ is the length scale parameter, $K_{\nu}$ is the modified Bessel function, and $\mathbf{GF}_i$, $\mathbf{GF}_j$ represent the geophysical feature vectors at locations $i$ and $j$. This formulation ensures that locations with similar geophysical features will have similar coefficients, which better captures the underlying physical relationships in Spatial-VCBN.

To allow for complex non-Gaussian distributions of causal coefficients, we transform the GP latent variables through a series of invertible normalizing flow transformations:

\begin{equation}
v^{l} = f_{K_{v}} \circ \cdots \circ f_2 \circ f_1(z_{v}^{l})
\label{NF_formula}
\end{equation}

\noindent where $f_1, ..., f_{K_{v}}$ are invertible transformations and $K_{v}$ is the number of flow layers. This combination allows us to model both spatial correlation (through the GP) and distributional complexity (through the normalizing flow).









For inference, we approximate the true posterior of the GP latent variables with a multivariate Gaussian variational distribution:
\begin{equation}
q(\mathbf{z}_{v}) = \mathcal{N}(\boldsymbol{\mu}_{v}, \boldsymbol{\Sigma}_{v})
\label{GPPosterior}
\end{equation}
\noindent where $\boldsymbol{\mu}_{v}$ represents the posterior mean vector and $\boldsymbol{\Sigma}_{v}$ the posterior covariance matrix across all locations. For computational efficiency, we parameterize $\boldsymbol{\Sigma}_{v}$ as a low-rank plus diagonal structure:
\begin{equation}
\boldsymbol{\Sigma}_{v} = \mathbf{L}_{v}(\mathbf{L}_{v})^T + \text{diag}(\boldsymbol{\delta}_{v})
\end{equation}
\noindent where $\mathbf{L}_{v}$ is a lower-triangular matrix of rank $r \ll n$ (with $n$ being the number of spatial locations), and $\boldsymbol{\delta}_{v}$ is a vector of positive diagonal elements.

To enable efficient gradient computation during optimization, we employ the reparameterization trick:
\begin{equation}
\mathbf{z}_{v} = \boldsymbol{\mu}_{v} + \mathbf{L}_{v}\boldsymbol{\epsilon}_1 + \text{diag}(\sqrt{\boldsymbol{\delta}_{v}})\boldsymbol{\epsilon}_2, \quad \boldsymbol{\epsilon}_1 \sim \mathcal{N}(\mathbf{0}, \mathbf{I}_r), \boldsymbol{\epsilon}_2 \sim \mathcal{N}(\mathbf{0}, \mathbf{I}_n)
\label{repara_GP}
\end{equation}
\noindent This allows us to sample from $q(\mathbf{z}_{v})$ while maintaining differentiability with respect to the variational parameters $\boldsymbol{\mu}_{v}$, $\mathbf{L}_{v}$, and $\boldsymbol{\delta}_{v}$.

When transforming the GP latent variables through the normalizing flow, we apply the change of variables formula to compute the density of the transformed variables:

\begin{equation}
\log q(v^{l} | GF) = \log q(z_{v}^{l}) - \sum_{k=1}^{K_{v}} \log |\det(\frac{\partial f_{k}}{\partial z_{v,k-1}^{l}})|
\label{change_of_variable_GP}
\end{equation}

\noindent where $z_{v,k-1}^{l}$ represents the output of the $(k-1)$-th transformation, with $z_{v,0}^{l} = z_{v}^{l}$.

Computing expectations under the transformed distribution is done via Monte Carlo sampling:

\begin{equation}
\mathbb{E}_{q(v^{l})}[h(v^{l})] = \mathbb{E}_{q(z_{v}^{l})}[h(f_{K_{v}} \circ \cdots \circ f_2 \circ f_1(z_{v}^{l}))]
\approx \frac{1}{M}\sum_{m=1}^{M}h(f_{K_{v}} \circ \cdots \circ f_2 \circ f_1(z_{v}^{l,(m)}))
\label{MC_GP}
\end{equation}

\noindent where $z_{v}^{l,(m)}$ are samples drawn from $q(z_{v}^{l})$ using the reparameterization in Equation \ref{repara_GP}, and $M$ is the number of Monte Carlo samples.

In our implementation, we utilize planar flows of the form:

\begin{equation}
f_k(z^{l}; u_{k}, c_{k}, b_{k}) = z^{l} + u_{k}h(c_{k}^T z^{l} + b_{k})
\end{equation}

\noindent where $h$ is a tanh activation function with derivative $h'(\cdot)$, and $u_{k}$, $c_{k}$, $b_{k}$ are learnable parameters of the $k$-th flow. The log-determinant of the Jacobian for this transformation is:

\begin{equation}
\log |\det(\frac{\text{d}f_k}{\text{d}z_{k-1}^{l}})| = \log |1 + u_{k}^T \psi(z_{k-1}^{l})|
\end{equation}

\noindent where $\psi(z) = c_{k} \cdot h'(c_{k}^T z + b_{k})$.



We parameterize the mean function $m_v(\mathbf{GF})$ as a function of the geophysical features:
\begin{equation}
m_v(\mathbf{GF}_l) = \text{NN}_v(\mathbf{GF}_l)
\end{equation}
\noindent where $\text{NN}_v$ is a neural network that maps geophysical features at location $i$ to the mean of the corresponding GP latent variable $z_v^l$.

This GP-Normalizing Flow approach provides several advantages for modeling spatially heterogeneous causal effects: (1) it naturally captures spatial correlation through the GP prior, (2) it allows for complex, non-Gaussian distributions through the normalizing flow transformation, and (3) it provides a principled way to incorporate uncertainty in the causal parameter estimates.

\subsection{Stochastic Variational Inference with Spatial-variant Causal Parameters}

The ultimate goal is to jointly infer the true posteriors of multiple unobserved target variables $x_i \in X$, which represents the target hazards and impacts with unknown parameters of causal dependencies, and the unknown spatial-varying causal parameters $\lambda_{a}^{l}, \text{where } a\in \{LS,LF, BD\}$, and $\gamma_{b}^{l}, \text{where } b \in \{\alpha_{LS}, \alpha_{LF}, LS, LF\}$, across different locations that quantify the causal dependencies among parent nodes and child nodes. Therefore, we use variational inference to approximate the true posteriors of $X^{l}$ using $q(X^{l})$ by optimizing the variational lower bound. Since the variable $X^{l}$ is binary, we can further factorize $q(X^{l})$ over hidden (unobserved) nodes by:

\begin{equation}
q(X^{l}) = \prod_{i\in\{LS, LF, BD\}}q(x_{i}^{l}) = \prod_{i\in\{LS, LF, BD\}}(q_{i}^{l})^{x_{i}^{l}}(1-q_{i}^{l})^{1-x_{i}^{l}}
\end{equation}

\noindent where $q_i^l$ is defined to approximate the posterior probability that node $i$ is active in location $l$.

We derive the lower bound of the marginal log-likelihood of the observed $Y$ by Jensen's inequality as follows:
\begin{equation}
\log P(Y, GF) = \underbrace{\mathbb{E}_{q(X^l,\epsilon^{l},\mathbf{z}^{l})}[\log p(y^{l}, GF, X^{l}, \mathbf{z}^{l})]}_{[1]} -  \underbrace{\mathbb{E}_{q(X^l,\epsilon^{l},\mathbf{z}^{l})}[\log q(X^l,\epsilon^{l},\mathbf{z}^{l})]}_{[2]}
\end{equation}
Here $\mathbf{z}^{l}$ represents the latent Gaussian Process variables, which are transformed through a normalizing flow to produce the spatially-varying parameters $\boldsymbol{\lambda}^{l}$ and $\boldsymbol{\gamma}^{l}$ shown in Equation \ref{NF_formula}.

The item [1] is further expanded as:
\begin{equation}
\begin{aligned}
\mathbb{E}_{q(X^l,\epsilon^{l},\mathbf{z}^{l})}[\log p(y^{l}, GF, X^{l}, \mathbf{z}^{l})] & = \underbrace{\mathbb{E}[\log p(y^{l} | x_{\mathcal{P}(y)}^{l}, 
f_{K_{\lambda}} \circ \cdots \circ f_2 \circ f_1 (\mathbf{z}^{l}_{\lambda}), \epsilon_{y}^{l}, GF)]}_{[3]} \\
 & + \underbrace{\sum_{i \in \{LS, LF, BD\}} \mathbb{E}[\log p(x_{i}^{l}|f_{K_{\gamma}} \circ \cdots \circ f_2 \circ f_1(\mathbf{z}^{l}_{\gamma \mathcal{P}(x_{i}^{l})}), \mathcal{P}(x^{l}), \epsilon_{i}^{l}, GF  ) ]}_{[4]}\\
 & + \underbrace{\mathbb{E}[\log p(\mathbf{z}^{l}_{\lambda}|GF)]}_{[5]}  + \underbrace{\mathbb{E}[\log p(\mathbf{z}^{l}_{\gamma}|GF)]}_{[6]}\\
& + \underbrace{\sum_{i \in \{LS, LF, BD\}} \mathbb{E}[\log p(\epsilon_{i}^{l})] +  \mathbb{E}[\log p(\epsilon_{y}^{l})]}_{C_{1}}
\label{[1]}
\end{aligned}
\end{equation}
Where $p(\mathbf{z}^{l}_{\lambda}|GF)$ and $p(\mathbf{z}^{l}_{\gamma}|GF)$ are Gaussian Process priors shown in Equation \ref{GPPrior}.
As for item [2], we have:
\begin{equation}
\begin{aligned}
\mathbb{E}_{q(X^l,\epsilon^{l},\mathbf{z}^{l})}[\log q(X^l,\epsilon^{l},\mathbf{z}^{l})] & = \underbrace{\sum_{i} \mathbb{E}(x_{i}^{l})\log q_{i}^{l} +  \sum_{i} \mathbb{E}(1- x_{i}^{l})\log (1-q_{i}^{l})}_{[7]}\\
&  + \underbrace{\mathbb{E}[\log q(\mathbf{z}^{l}_{\lambda})]}_{[8]}  + \underbrace{\mathbb{E}[\log q(\mathbf{z}^{l}_{\gamma})]}_{[9]}\\
& + \underbrace{\sum_{i} \mathbb{E}[\log q(\epsilon_{i}^{l})] +  \mathbb{E}[\log q(\epsilon_{y}^{l})]}_{C_{2}}
\label{[2]}
\end{aligned}
\end{equation}
where $C_1$ and $C_2$ cancel out. $q(\mathbf{z}^{l}_{\lambda})$ and $q(\mathbf{z}^{l}_{\gamma})$ are Gaussian Process posteriors presented in Equation \ref{GPPosterior}.

For terms [5], [6], [8], and [9] involving the Gaussian Process, we apply the KL divergence formula for multivariate Gaussians:
\begin{equation}
\begin{aligned}
[5] - [8] &= -KL[q(\mathbf{z}^{l}_{\lambda}) || p(\mathbf{z}^{l}_{\lambda}|\mathbf{GF})] \\
&= -\frac{1}{2}\left[\text{tr}(\mathbf{K}_{\lambda}^{-1}\boldsymbol{\Sigma}_{\lambda}) + (\boldsymbol{\mu}_{\lambda}-\mathbf{m}_{\lambda})^{T}\mathbf{K}_{\lambda}^{-1}(\boldsymbol{\mu}_{\lambda}-\mathbf{m}_{\lambda}) + \log|\mathbf{K}_{\lambda}| - \log|\boldsymbol{\Sigma}_{\lambda}| - n_{\lambda}\right]
\end{aligned}
\end{equation}

\begin{equation}
\begin{aligned}
[6] - [9] &= -KL[q(\mathbf{z}^{l}_{\gamma}) || p(\mathbf{z}^{l}_{\gamma}|\mathbf{GF})] \\
&= -\frac{1}{2}\left[\text{tr}(\mathbf{K}_{\gamma}^{-1}\boldsymbol{\Sigma}_{\gamma}) + (\boldsymbol{\mu}_{\gamma}-\mathbf{m}_{\gamma})^{T}\mathbf{K}_{\gamma}^{-1}(\boldsymbol{\mu}_{\gamma}-\mathbf{m}_{\gamma}) + \log|\mathbf{K}_{\gamma}| - \log|\boldsymbol{\Sigma}_{\gamma}| - n_{\gamma}\right]
\end{aligned}
\end{equation}

\noindent where $\mathbf{K}$ represents the prior covariance matrices derived from the Matern kernel over geophysical features, $\boldsymbol{\Sigma}$ denotes the variational posterior covariance matrices, $\boldsymbol{\mu}$ and $\mathbf{m}$ are the posterior and prior mean vectors respectively, and $n$ is the dimensionality of the corresponding latent variable vectors. These equations quantify the divergence between our variational approximation and the true GP prior distributions for both $\lambda$ and $\gamma$ coefficients.

With our assumption in the conditional distributions for $y^{l}|\mathcal{P}(y^{l}), \epsilon_{y}^{l}, \boldsymbol{\lambda_{\mathcal{P}(y^{l})}}^{l}$ shown in Equation \ref{logy_assumption}, we obtain the following. we can calculate item [3] in Equation \ref{[1]} as: 

\begin{equation}
\begin{aligned}
\mathbb{E} [\log p(y^{l}|\mathcal{P}(y^{l}), \epsilon_{y}^{l}, \boldsymbol{\lambda_{\mathcal{P}(y^{l})}}^{l})] & = - \log y^{l} - \log|w_{\epsilon_{y}}| - \frac{(\log y^{l}) + w_{0y}^{2} + \sum_{k\in\mathcal{P}(y^{l})} \mathbb{E}[(\lambda_{k}^{l})^{2}]q_{k}^{l} - 2\log y^{l}\cdot w_{0y}
 }{2w_{\epsilon_{y}}^{2}}\\
&  - \frac{\sum_{j\neq k} \mathbb{E}(\lambda_{k}^{l}) \cdot \mathbb{E}(\lambda_{j}^{l})\cdot q_{k}^{l}\cdot q_{j}^{l} + (w_{0y} - \log y^{l})(\sum_{k\in\mathcal{P}(y^{l})}\mathbb{E}(\lambda_{k}^{l})\cdot q_{k}^{l})}{w_{\epsilon_{y}}^{2}}
\label{Final_E_logpy}
\end{aligned}
\end{equation}

As for item [4] in Equation \ref{[1]}, we consider two scenarios - when $x_i$ is a leaf node and a non-leaf node. First we can describe the conditional distribution of LS, LF, and BD as follows:

\begin{equation}
\begin{aligned}
& p(x_{i}^{l}|x_{\mathcal{P}(x_{i})}^{l}, \epsilon_{i}^{l}, \boldsymbol{\gamma_{\mathcal{P}(x_{i}^{l})}}^{l}) = \\
& [\frac{1}{1+\exp(-\sum_{k \in \mathcal{P}(x_{i})}\gamma_{k}^{l} x_{k}^{l} - w_{\epsilon_{i}}\epsilon_{i}^{l} - w_{0i} )}]^{x_{i}^{l}}\cdot [\frac{1}{1+\exp(\sum_{k \in \mathcal{P}(x_{i})}\gamma_{k}^{l} x_{k}^{l} + w_{\epsilon_{i}}\epsilon_{i}^{l} + w_{0i} )}]^{1-x_{i}^{l}}
\end{aligned}
\label{logx}
\end{equation}

The expectation of Equation \ref{logx} can be formulated as:

\begin{equation}
\begin{aligned}
\mathbb{E}[\log p(x_{i}^{l}|x_{\mathcal{P}(x_{i})}^{l}, \epsilon_{i}^{l}, \boldsymbol{\gamma_{\mathcal{P}(x_{i}^{l})}}^{l})] & = q_{i}^{l} \mathbb{E}[-\log(1+\exp(-\sum_{k \in \mathcal{P}(x_{i})}\gamma_{k}^{l} x_{k}^{l} - w_{\epsilon_{i}}\epsilon_{i}^{l} - w_{0i} ))]\\
&  + (1- q_{i}^{l})\mathbb{E}[-\log(1+\exp(\sum_{k \in \mathcal{P}(x_{i})}\gamma_{k}^{l} x_{k}^{l} + w_{\epsilon_{i}}\epsilon_{i}^{l} + w_{0i} ))]
\end{aligned}
\label{E_log_px}
\end{equation}

However, the distribution of $-\log\left[1 + \exp\left(\sum_{k \in \mathcal{P}(x_i)} \gamma_k^l x_k^l + w_{\epsilon_i} \epsilon_i^l + w_{0i}\right)\right]$ is intractable, as it involves a log-sum-exp function that mixes both discrete and continuous variables. Consequently, we need to derive a tight lower bound for its expectation. Without loss of generality, we begin with the case where node $i$ has a single active parent. Given that the function $-\log x$ is convex, Jensen's inequality, combined with Taylor's theorem, allows us to establish the following relationship:

\begin{equation}
\mathbb{E}[-\log (1+\exp(x))]\ge -\log(1+\mathbb{E}[\exp(x)])
\end{equation}

Therefore, we obtain the lower bound of Equation \ref{E_log_px} as:

\begin{equation}
\begin{aligned}
\mathbb{E}[\log p(x_{i}^{l}|x_{\mathcal{P}(x_{i})}^{l}, \epsilon_{i}^{l}, \boldsymbol{\gamma_{\mathcal{P}(x_{i}^{l})}}^{l})] & \geq q_{i}^{l}[-\log(1+ \mathbb{E}[\exp(-\sum_{k \in \mathcal{P}(x_{i})}\gamma_{k}^{l} x_{k}^{l} - w_{\epsilon_{i}}\epsilon_{i}^{l} - w_{0i} )])]\\
& + (1-q_{i}^{l})[-\log(1+ \mathbb{E}[\exp(\sum_{k \in \mathcal{P}(x_{i})}\gamma_{k}^{l} x_{k}^{l} + w_{\epsilon_{i}}\epsilon_{i}^{l} + w_{0i} )])]\\
& = -q_{i}^{l}\Bigg\{\log\bigg[1 + \bigg(\prod_{k\in\mathcal{P}(x_{i})} \bigg[(1-q_{k}^{l}) + q_{k}^{l} \mathbb{E}[\exp(-\gamma_{k}^{l})]\bigg]\bigg)\cdot \exp(\frac{w_{\epsilon_{i}}^{2}}{2} - w_{0i}) \bigg]\Bigg \} \\
& - (1-q_{i}^{l})\Bigg\{\log\bigg[1 + \bigg(\prod_{k\in\mathcal{P}(x_{i})} \bigg[(1-q_{k}^{l}) + q_{k}^{l} \mathbb{E}[\exp(\gamma_{k}^{l})]\bigg]\bigg)\cdot \exp(\frac{w_{\epsilon_{i}}^{2}}{2} + w_{0i}) \bigg]\Bigg \}
\end{aligned}
\label{final_E_log_px}
\end{equation}

As for item [7] in Equation \ref{[2]}, we have:

\begin{equation}
\sum_{i} \mathbb{E}(x_{i}^{l})\log q_{i}^{l} +  \sum_{i} \mathbb{E}(1- x_{i}^{l})\log (1-q_{i}^{l}) = \sum_{i}  q_{i}^{l}\log q_{i}^{l} +  \sum_{i} ( 1-q_{i}^{l})\log (1-q_{i}^{l})
\end{equation}

Given a map containing a set of locations, $l \in L$, we further derive a tight lower bound for the log-likelihood as follows:
\begin{equation}
\begin{aligned}
\mathcal{L} = & \log P(Y, GF)\\
= &  - \log y^{l} - \log|w_{\epsilon_{y}}| - \frac{(\log y^{l}) + w_{0y}^{2} + \sum_{k\in\mathcal{P}(y^{l})} \mathbb{E}[(\lambda_{k}^{l})^{2}]q_{k}^{l} - 2\log y^{l}\cdot w_{0y}
 }{2w_{\epsilon_{y}}^{2}}\\
&  - \frac{\sum_{j\neq k} \mathbb{E}(\lambda_{k}^{l}) \cdot \mathbb{E}(\lambda_{j}^{l})\cdot q_{k}^{l}\cdot q_{j}^{l} + (w_{0y} - \log y^{l})(\sum_{k\in\mathcal{P}(y^{l})}\mathbb{E}(\lambda_{k}^{l})\cdot q_{k}^{l})}{w_{\epsilon_{y}}^{2}}\\
 & -\sum_{i\in\{LS, LF, BD\}} \Bigg[q_{i}^{l}\Bigg\{\log\bigg[1 + \bigg(\prod_{k\in\mathcal{P}(x_{i})} \bigg[(1-q_{k}^{l}) + q_{k}^{l} \mathbb{E}[\exp(-\gamma_{k}^{l})]\bigg]\bigg)\cdot \exp(\frac{w_{\epsilon_{i}}^{2}}{2} - w_{0i}) \bigg]\Bigg \} \\
& - (1-q_{i}^{l})\Bigg\{\log\bigg[1 + \bigg(\prod_{k\in\mathcal{P}(x_{i})} \bigg[(1-q_{k}^{l}) + q_{k}^{l} \mathbb{E}[\exp(\gamma_{k}^{l})]\bigg]\bigg)\cdot \exp(\frac{w_{\epsilon_{i}}^{2}}{2} + w_{0i}) \bigg]\Bigg \} \Bigg]\\
& -\sum_{a \in \{LS,LF,BD\}}\frac{1}{2}\left[\text{tr}(\mathbf{K}_{\lambda_{a}}^{-1}\boldsymbol{\Sigma}_{\lambda_{a}}) + (\boldsymbol{\mu}_{\lambda_{a}}-\mathbf{m}_{\lambda_{a}})^{T}\mathbf{K}_{\lambda_{a}}^{-1}(\boldsymbol{\mu}_{\lambda_{a}}-\mathbf{m}_{\lambda_{a}}) + \log|\mathbf{K}_{\lambda_{a}}| - \log|\boldsymbol{\Sigma}_{\lambda_{a}}| - n_{\lambda_{a}}\right]\\
& -\sum_{b \in \{\alpha_{LS}, \alpha_{LF}, LS, LF\}}\frac{1}{2}\left[\text{tr}(\mathbf{K}_{\gamma_{b}}^{-1}\boldsymbol{\Sigma}_{\gamma_{b}}) + (\boldsymbol{\mu}_{\gamma_{b}}-\mathbf{m}_{\gamma_{b}})^{T}\mathbf{K}_{\gamma_{b}}^{-1}(\boldsymbol{\mu}_{\gamma_{b}}-\mathbf{m}_{\gamma_{b}}) + \log|\mathbf{K}_{\gamma_{b}}| - \log|\boldsymbol{\Sigma}_{\gamma_{b}}| - n_{\gamma_{b}}\right]\\
& - \sum_{i}  q_{i}^{l}\log q_{i}^{l} -  \sum_{i} ( 1-q_{i}^{l})\log (1-q_{i}^{l})
\Bigg\}
\end{aligned}
\end{equation}

\subsection{Training and Stochastic Optimization with Sparse Gaussian Processes}

We aim to minimize our loss function in order to find optimal combinations of posteriors and parameters of causal dependencies (including the weights of parent nodes, and parameters of the normalizing flows). To achieve this, we develop an expectation-maximization (E--M) algorithm that alternates between updating the posteriors of unobserved variables (e.g., LS, LF, BD) with causal effects, and flow parameters.

To manage computational complexity when working with large geographical regions, we employ a sparse Gaussian Process formulation using inducing points. Rather than modeling the GP latent field directly at every location, we introduce a set of $M$ inducing points $\mathbf{u}_v$ at strategic locations in feature space. This approach reduces the computational complexity from $O(N^3)$ to $O(NM^2)$, where $N$ is the number of locations and $M \ll N$ is the number of inducing points.

We model spatial correlation in the causal parameter field using a Gaussian Process with a Matern kernel operating in feature space rather than geographical space. This approach allows the model to capture correlations between locations with similar geophysical characteristics (elevation, slope, lithology, etc.) even when they are not physically adjacent. This is motivated by the observation that similar feature combinations tend to exhibit similar causal relationships in earthquake-triggered hazard scenarios regardless of their geographic proximity.

Within each iteration, we sample a mini-batch of locations and perform the following two steps:

\begin{itemize}
    \item \textbf{Expectation step:} Update the posterior probability estimates of the unobserved latent variables (LS, LF, BD) at each location by conditioning on the current GP parameters, neural network weights, and the most recent samples of causal coefficients derived from the GP latent variables through normalizing flows. The posterior probabilities are updated according to the causal dependencies encoded in the Bayesian network structure.
    
    \item \textbf{Maximization step:} Update the parameters of the sparse Gaussian Process, specifically the variational parameters for inducing points $\mathbf{u}_v$, along with the neural network parameters that map geophysical features to the mean function of the GP. Specifically, for iteration $t + 1$, these parameters are updated as
    \begin{equation}
        \boldsymbol{\mu}_u^{(t+1)} = \boldsymbol{\mu}_u^{(t)} + \rho \mathbf{A}\,\nabla \mathcal{L}(\boldsymbol{\mu}_u), \quad 
        \boldsymbol{\Sigma}_u^{(t+1)} = \boldsymbol{\Sigma}_u^{(t)} + \rho \mathbf{A}\,\nabla \mathcal{L}(\boldsymbol{\Sigma}_u),
    \end{equation}
    \begin{equation}
        \boldsymbol{\theta}_{NN}^{(t+1)} = \boldsymbol{\theta}_{NN}^{(t)} + \rho \mathbf{A}\,\nabla \mathcal{L}(\boldsymbol{\theta}_{NN}),
    \end{equation}
    where $\boldsymbol{\mu}_u$ and $\boldsymbol{\Sigma}_u$ are the parameters of the variational distributions over inducing points, $\boldsymbol{\theta}_{NN}$ represents the neural network parameters, $\mathbf{A}$ is a positive definite preconditioner, $\rho$ is the learning rate, and $\nabla \mathcal{L}(\cdot)$ denotes gradients of the loss function with respect to the corresponding parameters. The causal coefficients at observed locations are then derived through conditioning on the inducing points and applying normalizing flow transformations to the resulting GP latent variables. This gradient update scheme is guaranteed to converge to a local maximum of $\mathcal{L}$ if $\rho$ satisfies appropriate decay conditions.
\end{itemize}

Our variational evidence lower bound (ELBO) with inducing points can be formulated as:
\begin{equation}
\mathcal{L} = \mathbb{E}_q[\log p(\mathbf{y}|\mathbf{X},\mathbf{z}_v)] + \mathbb{E}_q[\log p(\mathbf{X}|\mathbf{z}_v)] - KL[q(\mathbf{u}_v)||p(\mathbf{u}_v)] - \mathbb{E}_q[\log q(\mathbf{X})]
\end{equation}

where $q(\mathbf{X},\mathbf{z}_v,\mathbf{u}_v,\boldsymbol{\epsilon}) = q(\mathbf{X}) \times p(\mathbf{z}_v|\mathbf{u}_v) \times q(\mathbf{u}_v) \times p(\boldsymbol{\epsilon})$ is our variational distribution. By using the exact conditional $p(\mathbf{z}_v|\mathbf{u}_v)$ in our variational approximation, we ensure that the model maintains valid probabilistic semantics over the entire spatial field while only needing to represent distributions at the inducing points explicitly.

Once the model converges, we obtain the final posterior estimates of LS, LF, and BD for each location, and also the causal effects between hidden hazards and observations, reflecting the spatially heterogeneous causal relationships inferred from the observed data.

We also apply a local pruning algorithm to accelerate the computation over a large region. This strategy is motivated by the observation that real-world causal graphs are typically sparse: only a small subset of nodes stay active. For example, locations without building footprints will not have damaged buildings, i.e., building damage nodes are inactive. Therefore, we can prune these inactive nodes while keeping the active ones crucial for parameter updates \cite{xu2022seismic,li2023disasternet}.

\section{Results}\label{sec2}

\subsection{Data Description}
We evaluate our framework on three earthquake events: the 2020 Puerto Rico earthquake (M6.4), 2021 Haiti earthquake (M7.2), and 2023 Turkey-Syria earthquake sequence (M7.8).

\noindent\textbf{The 2020 Puerto Rico earthquake}
A magnitude 6.4 earthquake hit the southwest part of Puerto Rico on January 7, 2020. The ARIA team created damage proxy maps using SAR images from the Sentinel-1 satellite to identify potentially damaged areas \cite{aria2020pr}. Researchers from the USGS, the University of Puerto Rico Mayag\"uez, the GEER team, and the StEER team later conducted field reconnaissance to collect ground truth observations \cite{allstadt2022ground,miranda2020,miranda2020b}. Post-earthquake reports documented that at least 300 landslides were triggered near the epicenter \cite{allstadt2022ground}.

\noindent\textbf{The 2021 Haiti earthquake}
On August 14, 2021, a magnitude 7.2 earthquake struck the southern peninsula of Haiti. The StEER team and GEER team later collected ground truth inventories for landslides and building damage \cite{zhao2022evaluation,HaitiBDGT,HaitiGEER}. According to post-disaster reports, the earthquake resulted in at least 2,248 human fatalities, destroyed 53,815 buildings, and damaged 83,770 structures throughout Grand Anse, Nippes and Sud \cite{reportHaiti,priorHaiti}.

\noindent\textbf{The 2023 Turkey-Syria earthquake sequence}
On February 6, 2023, an earthquake of Mw 7.8 and its aftershocks caused unparalleled destruction in Turkey and Syria. This disaster resulted in more than 55,000 deaths, displaced 3 million people in Turkey and 2.9 million in Syria, and caused damage or destruction to at least 230,000 buildings. \cite{li2025rapid, li2023m7}. The ARIA team generated DPM derived from synthetic aperture radar (SAR) images on Feb. 10, 2023 by the Copernicus Sentinel-1 satellites operated by the European Space Agency (ESA) \cite{ARIA1}. Ground truth was later collected and reported by the Turkish Ministry of Environment \cite{turkeyadm, li2025rapid}.

\subsection{Evaluation Metrics and Benchmarks}

We evaluate our framework using multiple complementary metrics to provide a comprehensive assessment of performance. The primary evaluation is based on the receiver operating characteristics (ROC) curve and its area under curve (AUC) metric \cite{fawcett2006introduction}. ROC curves plot True Positive Rate (TPR) against False Positive Rate (FPR) across varying decision thresholds, providing a threshold-independent assessment that is particularly valuable in disaster contexts where optimal classification thresholds may vary by event type, geographic region, or hazard category. This approach aligns well with the probabilistic nature of our framework, as both our spatially-aware causal model and the benchmark methods produce confidence scores rather than hard classifications.

To complement the AUC metric, we also employ the F1 score, which is the harmonic mean of precision and recall. The F1 score is calculated as $F1 = 2 \times \frac{precision \times recall}{precision + recall}$, where precision represents the ratio of correctly identified hazards to all predicted hazards, and recall captures the proportion of actual hazards that were correctly identified. Unlike AUC, which evaluates performance across all possible thresholds, F1 score requires selecting a specific classification threshold. For consistent comparison, we use the threshold that maximizes the F1 score for each model and hazard type. This metric is particularly informative for disaster response applications, where balancing false alarms (precision) with missed hazards (recall) is critical for effective resource allocation.

Although our framework is fundamentally an unsupervised model that approximates true posteriors through variational inference, these evaluation metrics enable objective comparison with supervised approaches. Additionally, we utilize the variational bound as an internal metric to optimize model parameters and determine the optimal flow length for the normalizing flow component.

\subsection{Spatially-Varying Causal Influence Mapping between Multi-Hazards and Structural Damage}

\begin{figure}[htbp]
\begin{center}
\scalebox{1}{
\begin{tabular}{c}
\includegraphics[width=1\columnwidth]{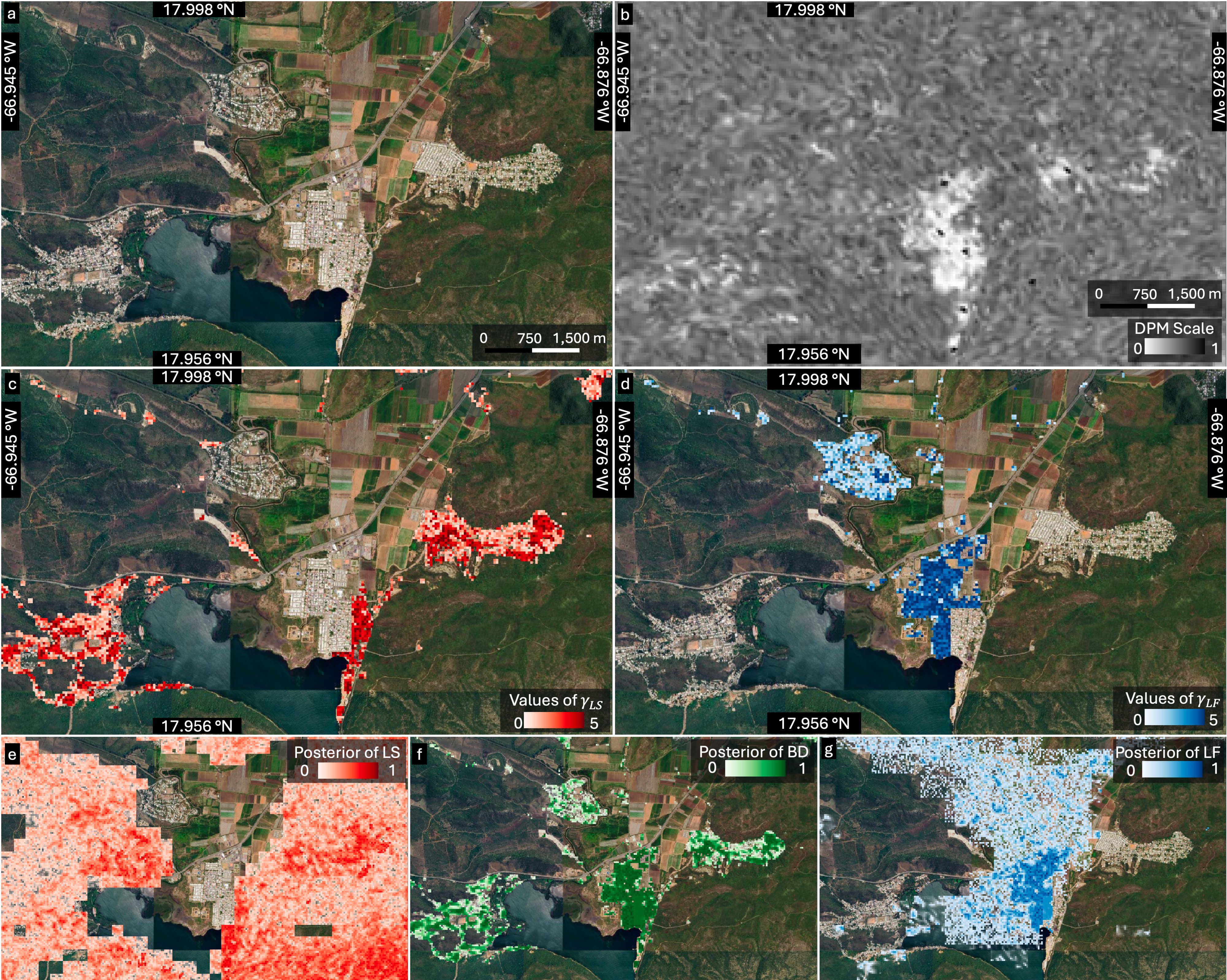}
\end{tabular}}
\caption{Spatial distribution of hazards, causal parameters, and damage following the 2020 Puerto Rico earthquake. (a) Google satellite imagery of the study area with extent of $-66.945 \degree W$, $17.956 \degree N$ to $-66.876 \degree W$, $17.998 \degree N$. (b) DPM derived from satellite imagery. (c) Spatial distribution of $\gamma_{LS}$ values, showing the causal parameter strength from landslides (LS) to building damage (BD). (d) Spatial distribution of $\gamma_{LF}$ values, showing the causal parameter strength from liquefaction (LF) to building damage (BD). (e) Posterior probability of landslide occurrence. (f) Posterior probability of building damage occurrence. (g) Posterior probability of liquefaction occurrence.}
\label{PR_gamma_posteriors}
\vspace{-0.5cm}
\end{center}
\end{figure}

Figure \ref{PR_gamma_posteriors} illustrates the complex spatial relationships between different types of hazard, their causal influence, and surface deformation patterns following the 2020 Puerto Rico earthquake with an extent of $-66.945 \degree W$, $17.956 \degree N$ to $-66.876 \degree W$, $17.998 \degree N$. The DPM (Figure \ref{PR_gamma_posteriors}(b)) captures surface deformation and ground changes, serving as an observable proxy that Spatial-VCBN interprets through causal relationships.

The spatial distribution of the causal parameters $\gamma_{LS}$  and $\gamma_{LF}$, which are shown in Figures \ref{PR_gamma_posteriors}(c)(d), reveals how different hazard mechanisms contribute to the damage of buildings throughout the landscape. Areas with elevated $\gamma_{LS}$ values indicate locations where landslides have a stronger causal influence on building damage, predominantly in the hillier portions of the study area. These high-coefficient regions align well with areas showing a high posterior probability of landslide occurrence (Figure \ref{PR_gamma_posteriors}(e)), suggesting that Spatial-VCBN successfully captures the spatial specificity of landslide impacts.

The effects of liquefaction show distinctly different spatial patterns, with high values of $\gamma_{LF}$ concentrated in the central lowlands and coastal areas. This clear spatial separation between high $\gamma_{LS}$ and high $\gamma_{LF}$ regions validates Spatial-VCBN assumption that landslides and liquefaction generally do not co-occur at the same location due to their different geological requirements. The posterior probability of building damage (Figure \ref{PR_gamma_posteriors}(f)) represents the result of our causal network, showing how the model integrates information from both types of hazard. Importantly, the model does not simply translate DPM signals directly into damage estimates but interprets these signals through the learned causal structure and spatially varying parameters.

These results demonstrate the ability of Spatial-VCBN to decouple different causal mechanisms contributing to observed surface changes, providing a more nuanced understanding of disaster impacts than would be possible from DPM interpretation alone. This capability forms the foundation for our subsequent analysis of how these causal parameters influence remote sensing observations.

Building on our understanding of causal relationships between hazards and structural damage, we now examine how these mechanisms present in DPM signals under varying conditions. We analyze both high-fidelity signal regions and challenging signal-constrained environments to demonstrate the robustness of Spatial-VCBN.

\subsection{Robust Spatially-Varying Causal Inference Between Hazards and Remote Sensing Observations}

\begin{figure}[t]
\begin{center}
\scalebox{1}{
\begin{tabular}{c}
\includegraphics[width=1\columnwidth]{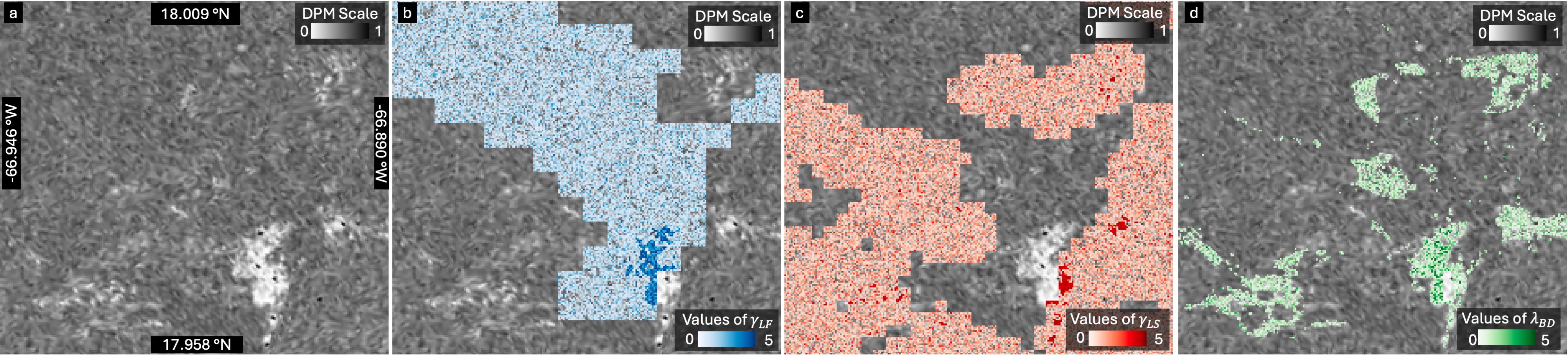}
\end{tabular}}
\caption{Spatial distribution of causal parameters and DPM signals in a high-deformation area. (a) Damage Proxy Map (DPM) showing surface deformation with brighter areas indicating higher change detection values in the study area with extent of $-66.946 \degree W$, $17.958 \degree N$ to $-66.890 \degree W$, $18.009 \degree N$. (b) Spatial distribution of $\lambda_{LF}$, quantifying the causal relationship strength from liquefaction to DPM signals. (c) Spatial distribution of $\lambda_{LS}$, quantifying the causal relationship strength from landslides to DPM signals. (d) Spatial distribution of $\lambda_{BD}$, quantifying the causal relationship strength from building damage to DPM signals.}
\label{PRLSLFBDlambda}
\vspace{-0.5cm}
\end{center}
\end{figure}

Figure \ref{PRLSLFBDlambda} reveals the complex spatial relationships between different hazard mechanisms and their contributions to the observed DPM signals in an area with significant surface deformation in the study area with an extent of $-66.946 \degree W$, $17.958 \degree N$ to $-66.890 \degree W$, $18.009 \degree N$. The DPM presented in Figure \ref{PRLSLFBDlambda}(a) shows a concentrated area with high deformation values. The spatial distribution of the causal parameters provides insight into the primary mechanisms that drive these changes.

In this region, we observe spatial heterogeneity in how different hazards influence the DPM signals. The liquefaction causal parameter $\lambda_{LF}$ (Figure \ref{PRLSLFBDlambda}(b)) shows stronger values in the lower portion of the deformation area, suggesting that liquefaction processes are a dominant contributor to the DPM signals in that specific zone. This pattern is consistent with the typical geographic distribution of liquefaction in low-lying areas with specific soil conditions. Conversely, the landslide causal parameter $\lambda_{LS}$ shown in Figure \ref{PRLSLFBDlambda}(c) exhibits higher values in scattered patches, particularly in the right portion of the image where several high-intensity areas appear. This indicates locations where landslide mechanisms have a stronger influence on the observed DPM signals, likely corresponding to areas with steeper slopes or unstable terrain.

Figure \ref{PRLSLFBDlambda}(d) displays the building damage parameter $\lambda_{BD}$. It shows a distinct spatial pattern with moderate to high values in several discrete clusters. These areas represent locations where building deformation and structural impacts most strongly contribute to the DPM signals, which often correspond to zones with higher density of built structures. This spatial segregation of causal parameters demonstrates the ability of Spatial-VCBN to decouple multiple contributing factors to DPM signals, providing insights into the dominant hazard mechanisms at different locations within the affected area.

This capability to decouple in high-signal regions establishes a baseline for comparison with more challenging signal-constrained environments, which we examine next to demonstrate the robustness of Spatial-VCBN under varying conditions.

\begin{figure}[t]
\begin{center}
\scalebox{1}{
\begin{tabular}{c}
\includegraphics[width=1\columnwidth]{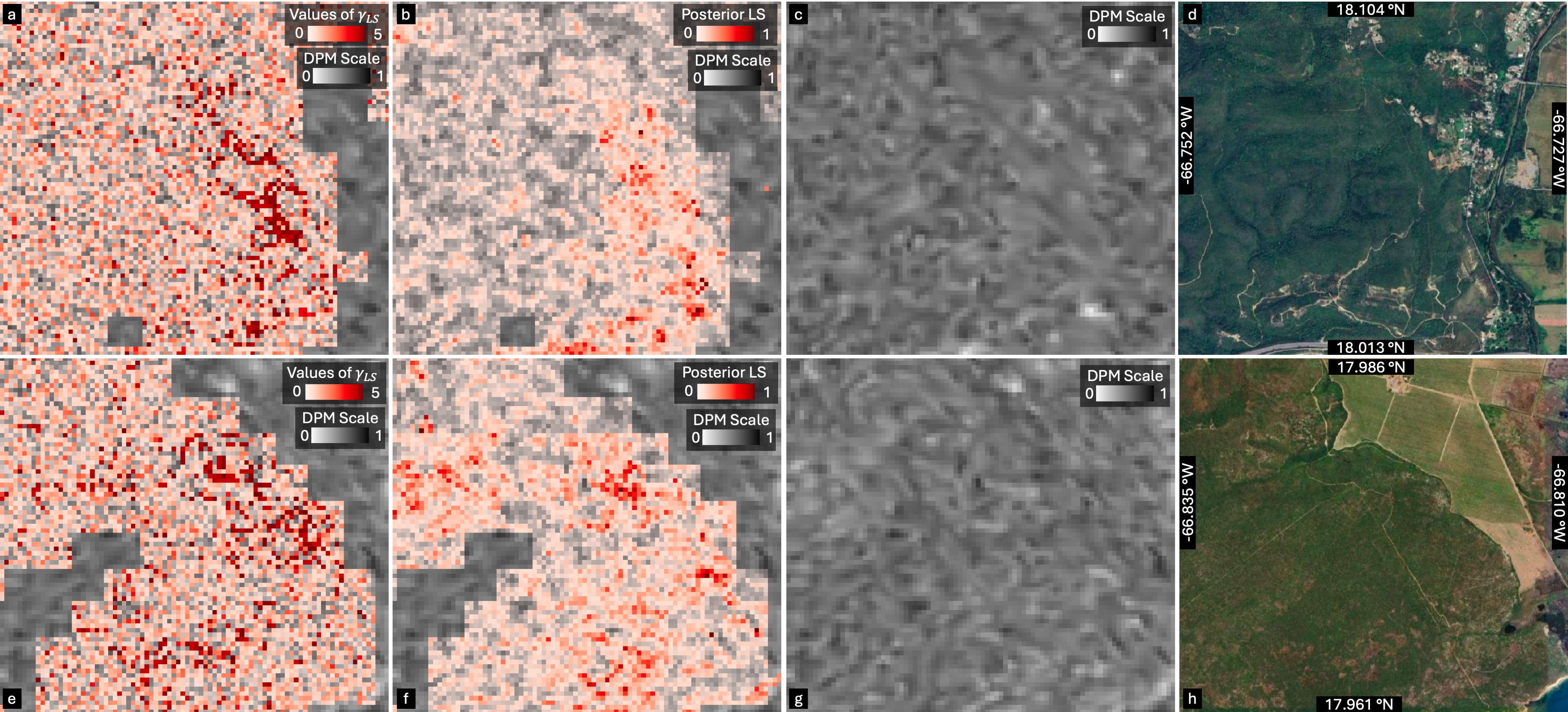}
\end{tabular}}
\caption{Spatial pattern learning in areas with noisy DPM signals following the 2020 Puerto Rico earthquake. This figure demonstrates the ability of Spatial-VCBN to detect landslide patterns even in areas with weak or noisy DPM signals. Top row (a-d) and bottom row (e-h) show two different regions following the 2020 Puerto Rico earthquake: (a,e) Spatial distribution of $\lambda_{LS}$, quantifying the causal relationship strength from landslides to DPM; (b,f) Posterior probability of landslide occurrence; (c,g) Damage Proxy Map (DPM) showing noisy surface deformation signals (0-1); (d,h) Google satellite imagery showing the corresponding terrain.}
\label{PRLSlambda}
\vspace{-0.5cm}
\end{center}
\end{figure}

While Spatial-VCBN performs well in areas with strong DPM signals, real-world disaster assessment often involves regions with weak, noisy, or ambiguous observational data. We now demonstrate the ability of Spatial-VCBN to extract meaningful causal parameters even in these challenging signal-constrained environments. Figure \ref{PRLSlambda} highlights the robust capability of Spatial-VCBN to identify meaningful landslide patterns even in challenging conditions where DPM signals are weak, inconsistent, or contaminated by noise. The areas shown exhibit noisy DPM readings, shown in Figures \ref{PRLSlambda}(c)(g) that likely result from environmental factors such as snow cover or the complex reflectance properties of mountainous terrain, which can introduce artifacts in satellite-based change detection. Despite these challenging conditions, our spatially-aware causal Bayesian network successfully extracts coherent landslide susceptibility patterns. The distributions of landslide causal parameter $\lambda_{LS}$ shown in Figures \ref{PRLSlambda}(a)(e) display distinct spatial organization with higher values concentrated along features that correspond to terrain characteristics associated with landslide risk. This demonstrates that Spatial-VCBN can effectively filter signal from noise by leveraging spatial correlation structures through its Gaussian Process component.

The posterior probability maps for landslide occurrence presented in Figures \ref{PRLSlambda}(b)(f) reveal a refined understanding of landslide risk that transcends the limitations of the noisy DPM data. These posterior estimates incorporate both the learned causal strengths and the underlying geophysical context, resulting in coherent spatial patterns that align with landslide-prone landscape features visible in the satellite imagery (Figures \ref{PRLSlambda}(d)(h)). This ability to maintain signal fidelity in challenging conditions is particularly important for comprehensive hazard assessment in mountainous regions, where environmental factors typically confound traditional analysis methods.

\begin{figure}[t]
\begin{center}
\scalebox{1}{
\begin{tabular}{c}
\includegraphics[width=1\columnwidth]{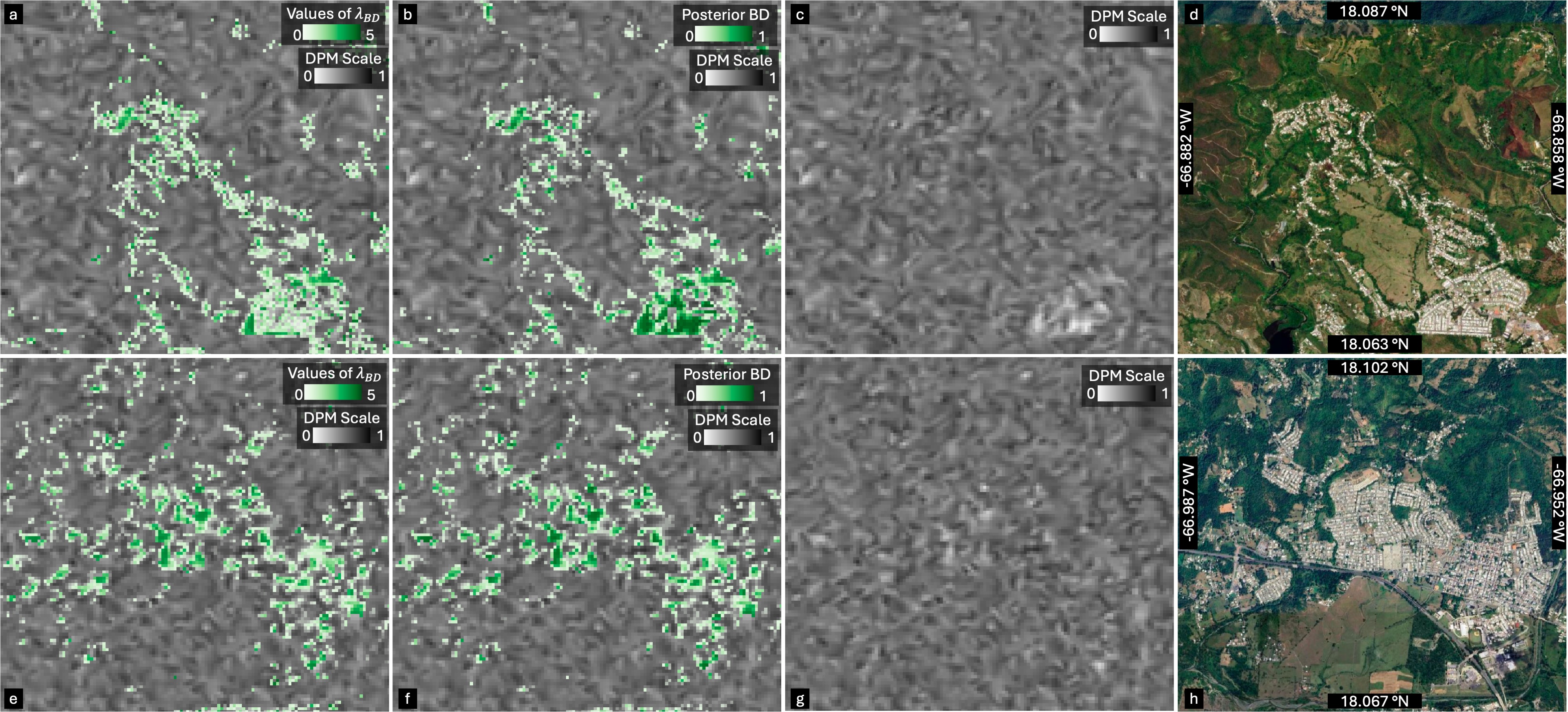}
\end{tabular}}
\caption{Building damage estimation in areas with noisy DPM signals following the 2020 Puerto Rico earthquake. This figure demonstrates the ability of Spatial-VCBN to detect building damage patterns despite weak DPM signals. Top row (a-d) and bottom row (e-h) show two different regions. (a,e) Spatial distribution of $\lambda_{BD}$, quantifying the causal relationship strength from building damage to DPM; (b,f) Posterior probability of building damage occurrence; (c,g) DPM showing noisy surface deformation signals; (d,h) Google satellite imagery showing developed areas with building structures.}
\label{PRBDlambda}
\vspace{-0.5cm}
\end{center}
\end{figure}

Similar resilience is observed in building damage estimation, as shown in Figure \ref{PRBDlambda}. Several factors can introduce noise in building damage estimation from satellite-based DPM, including variable building materials and construction types that respond differently to deformation, vegetation coverage partially obscuring buildings, complex urban geometries creating shadows and radar reflection artifacts, temporal variations in atmospheric conditions affecting satellite measurements, and pre-existing structural modifications unrelated to disaster impacts. Despite these challenges, our spatially-aware causal Bayesian network successfully extracts coherent building damage patterns. The distributions of building damage causal parameter $\lambda_{BD}$ shown in Figures \ref{PRBDlambda}(a)(e) exhibit clear spatial structure with higher values concentrated along what appear to be developed corridors visible in the satellite imagery. These patterns follow the distribution of built environments rather than appearing randomly distributed, suggesting that Spatial-VCBN is able to capture true building-related signals despite the noise.

The posterior probability maps for building damage presented in Figures \ref{PRBDlambda}(b)(f) show improved spatial patterns that are more structured than what might be inferred from the noisy DPM alone. The larger posterior probabilities appear primarily in areas with visible building clusters in the satellite imagery (Figures \ref{PRBDlambda}(d)(h)), demonstrating that Spatial-VCBN effectively incorporates contextual information about the built environment.

\begin{figure}[t]
\begin{center}
\scalebox{1}{
\begin{tabular}{c}
\includegraphics[width=1\columnwidth]{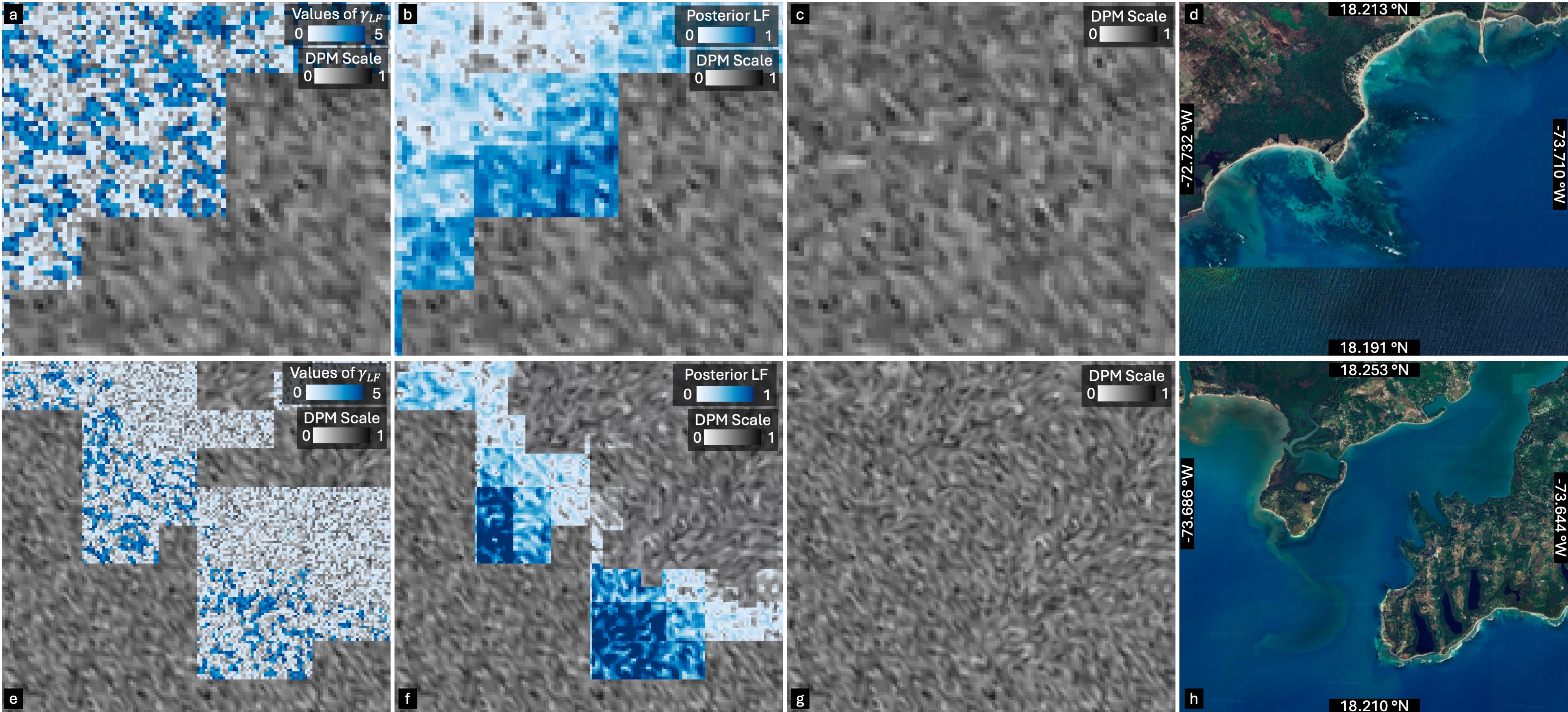}
\end{tabular}}
\caption{Liquefaction estimation in coastal areas of the 2021 Haiti earthquake. Top row (a-d) and bottom row (e-h) show two different coastal areas. (a,e) Spatial distribution of $\lambda_{LF}$, quantifying the causal relationship strength from liquefaction to DPM; (b,f) Posterior probability of liquefaction occurrence; (c,g) DPM showing noisy surface deformation signals; (d,h) Google satellite imagery showing coastal areas susceptible to liquefaction.}
\label{HTLFlambda}
\vspace{-0.5cm}
\end{center}
\end{figure}

The robustness of Spatial-VCBN extends to different earthquake events and hazard types, as demonstrated in Figure \ref{HTLFlambda} for liquefaction estimation in the 2021 Haiti earthquake. Even when DPM signals are compromised by various confounding factors such as water level fluctuations, coastal erosion processes, wave action affecting shoreline appearance, soil moisture variations in near-shore environments, and varying sediment compositions that respond differently to seismic shaking, Spatial-VCBN successfully identifies coherent liquefaction patterns across different coastal areas. The distributions of liquefaction causal parameter $\lambda_{LF}$, which is presented in Figures \ref{HTLFlambda}(a)(e), show distinctive spatial patterns with higher values concentrated in areas near coastlines where geological conditions favor liquefaction. The consistency of these patterns across different coastal regions within the same earthquake event suggests that Spatial-VCBN is able to capture fundamental physical relationships rather than location-specific anomalies.

The posterior probability maps for liquefaction, shown in Figures \ref{HTLFlambda}(b)(f), exhibit refined spatial patterns that align with coastal geomorphology visible in the satellite imagery (Figures \ref{HTLFlambda}(d)(h)). Higher probabilities appear in low-lying coastal areas with likely unconsolidated sediments that are typically more susceptible to liquefaction during seismic events. This correspondence between model predictions and physical geography further validates the ability of Spatial-VCBN to capture meaningful hazard patterns despite noisy observations.

These qualitative assessments across different hazard types, signal conditions, and earthquake events demonstrate the robust nature of our spatially-aware causal framework. In the following section, we provide quantitative evaluation metrics that further validate these observations.

Figure \ref{HTLFlambda} demonstrates the effectiveness of Spatial-VCBN in detecting liquefaction patterns across different coastal regions affected by the 2021 Haiti earthquake, even when DPM signals are compromised by various confounding factors such as water level fluctuations, coastal erosion processes, wave action affecting shoreline appearance, soil moisture variations in near-shore environments, and varying sediment compositions that respond differently to seismic shaking. Despite these challenges, our spatially-aware causal Bayesian network successfully identifies coherent liquefaction patterns across different coastal areas. The distributions of liquefaction causal parameter $\lambda_{LF}$, which is presented in Figures \ref{HTLFlambda}(a)(e), show distinctive spatial patterns with higher values concentrated in areas near coastlines where geological conditions favor liquefaction. The consistency of these patterns across different coastal regions within the same earthquake event suggests that Spatial-VCBN is able to capture fundamental physical relationships rather than location-specific anomalies. The posterior probability maps for liquefaction, shown in Figures \ref{HTLFlambda}(b)(f), exhibit refined spatial patterns that align with coastal geomorphology visible in the satellite imagery (Figures \ref{HTLFlambda}(d)(h)). Higher probabilities appear in low-lying coastal areas with likely unconsolidated sediments that are typically more susceptible to liquefaction during seismic events. This correspondence between model predictions and physical geography further validates the ability of Spatial-VCBN to capture meaningful hazard patterns despite noisy observations. The ability of Spatial-VCBN to produce consistent, physically plausible liquefaction assessments in challenging coastal environments, where traditional DPM analysis would be heavily compromised by noise, demonstrates its value for comprehensive multi-hazard assessment in diverse settings.

\subsection{Cross-Event Validation and Baseline Comparison}

\begin{figure}[t]
\begin{center}
\scalebox{1}{
\begin{tabular}{c}
\includegraphics[width=1\columnwidth]{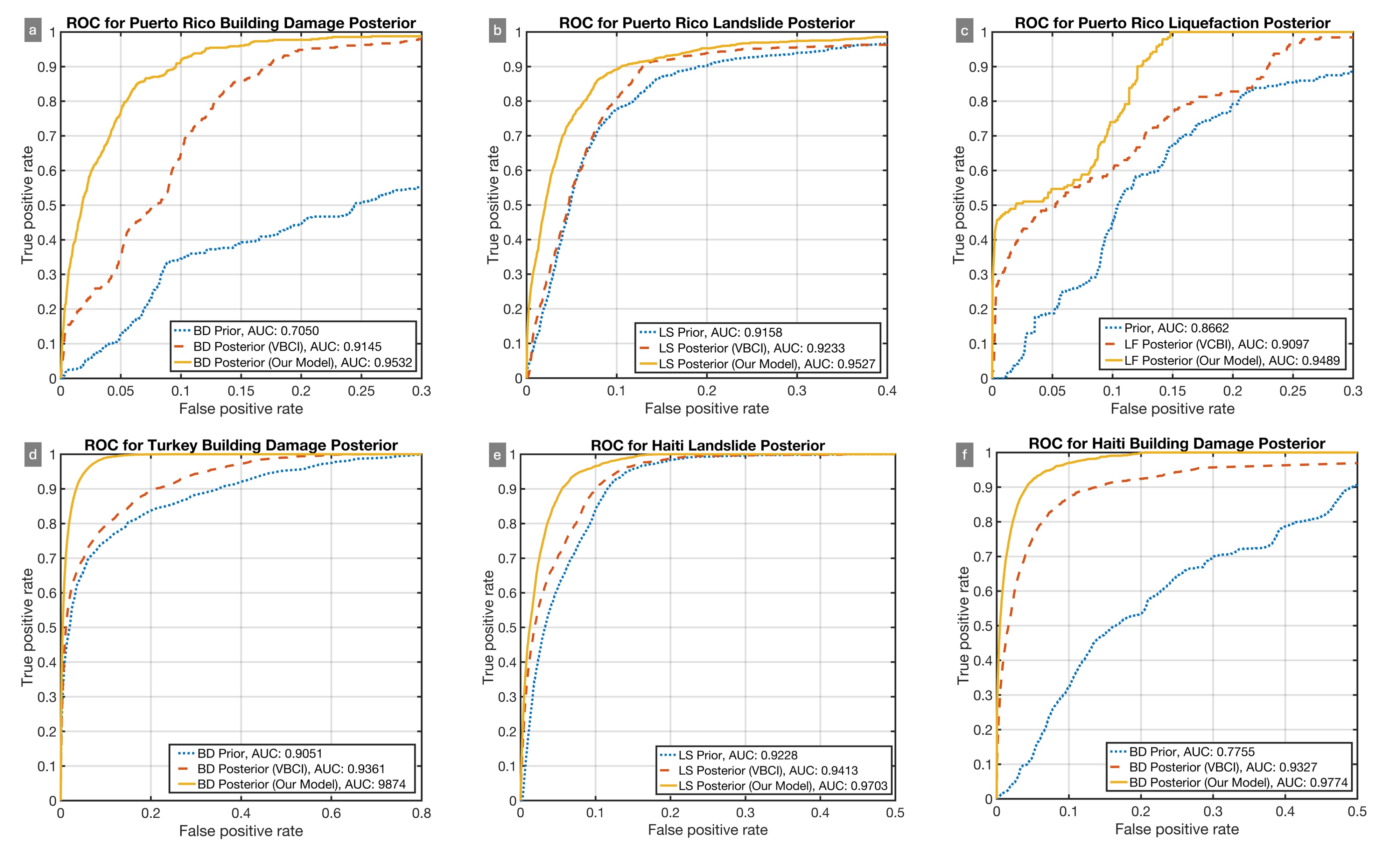}
\end{tabular}}
\caption{ROC (Receiver Operating Characteristic) curves and AUC (Area Under the Curve) values comparing the performance of Spatial-VCBN against baselines for different hazard types across three recent earthquakes. (a-c) Building damage, Landslide, and Liquefaction ROC curves of the 2020 Puerto Rico earthquake.  (d) Building damage for 2023 Turkey-Syria earthquake, (e-f) Building damage and Landslide ROC curves of the 2021 Haiti earthquake. Each plot compares three approaches: prior probability \cite{zhu2016updated,nowicki2018global} (blue dotted line), posterior from VBCI baseline \cite{xu2022seismic} (red dashed line), and posterior from Spatial-VCBN (solid yellow line).}
\label{ROC}
\vspace{-0.5cm}
\end{center}
\end{figure}

Our evaluation framework includes a comprehensive array of comparative methods across multiple hazard types. For our landslide (LS) and liquefaction (LF) estimation performance, we evaluate against several established approaches: the ground failure models developed by the USGS \cite{zhu2016updated,nowicki2018global}, the VBCI methodology proposed by \cite{xu2022seismic}, as well as both the Artificial Neural Network (ANN) and Gradient Boosting Machine (GBM) techniques outlined by \cite{novellino2021slow}. In assessing building damage (BD) estimation capabilities, we contrast Spatial-VCBN with conventional building fragility curves, the VBCI framework \cite{xu2022seismic}, ensemble methodology \cite{rao2023earthquake}, and the bilateral filtering approach recently introduced by \cite{li2024spatial}. Figure \ref{ROC} and Tables \ref{ROCrealCombined} and \ref{F1realCombined} demonstrate the superior performance of our spatially-aware causal Bayesian network across multiple hazard types and earthquake events. The consistently higher metric values achieved by Spatial-VCBN validate the effectiveness of Spatial-VCBN in accurately identifying disaster impacts.

For the 2020 Puerto Rico earthquake, Spatial-VCBN achieves remarkable improvements in building damage (Figure \ref{ROC}(a)), with an AUC of 0.9532 compared to 0.9145 for the VBCI and only 0.7050 for the prior. This represents a substantial 35.2\% improvement over the prior probability baseline. As shown in Table \ref{ROCrealCombined}, Spatial-VCBN consistently outperforms six competitive baselines, including both traditional statistical approaches and deep learning methods. The F1 score results shown in Table \ref{F1realCombined} further validate these findings, with Spatial-VCBN achieving 0.9167 for building damage in Puerto Rico, compared to the next best performer (Bilateral filter \cite{li2024spatial}) at 0.8945. This performance gain is particularly noteworthy for building damage, where distinguishing between different damage mechanisms can be challenging. For landslides (Figure \ref{ROC}(b)), Spatial-VCBN also outperforms competitors with an AUC of 0.9527, though the margin is smaller as all methods perform relatively well on this hazard type. Looking at Table \ref{ROCrealCombined}, we observe that the closest competitor for landslide estimation is the Bilateral filter approach (0.9413), which still falls short of the performance of  our method by 1.2\%. The consistency between AUC and F1 metrics (0.9173 for Spatial-VCBN versus 0.9031 for Bilateral filter) indicates that our performance improvements are robust across different evaluation criteria. Liquefaction estimation (Figure \ref{ROC}(c)) shows intermediate improvements, with Spatial-VCBN achieving an AUC of 0.9489, representing a 9.5\% improvement over the prior (0.8662). Table \ref{F1realCombined} shows a similar pattern for F1 scores, with Spatial-VCBN achieving 0.9107 compared to 0.8815 for Bilateral filter and 0.8714 for VBCI, highlighting the balanced performance of Spatial-VCBN in both estimation accuracy and false alarm reduction.

\begin{table*}[t]
\begin{center}
\caption{\textbf{AUC values comparison across earthquake events.} LS: Landslide, LF: Liquefaction, BD: Building Damage. HT: Haiti, PR: Puerto Rico, TK: Turkey. NGA: No Ground Truth Available. Bold values represent the best performance for each column.}
\begin{tabular}{l|ccc|ccc|ccc}
\toprule
\multirow{2}{*}{\textbf{Model}} & \multicolumn{3}{c|}{\textbf{Landslides}} & \multicolumn{3}{c|}{\textbf{Liquefaction}} & \multicolumn{3}{c}{\textbf{Building Damage}} \\
& HT & PR & TK & HT & PR & TK & HT & PR & TK \\
\midrule
\textbf{Spatial-VCBN} & \textbf{0.9703} & \textbf{0.9527} & NGA & NGA & \textbf{0.9489} & NGA & \textbf{0.9774} & \textbf{0.9532} & \textbf{0.9674} \\
Prior Model \cite{zhu2016updated,nowicki2018global,fema2003hazus} & 0.9228 & 0.9158 & NGA & NGA & 0.8662 & NGA & 0.7755 & 0.7050 &  0.9051\\
VBCI \cite{xu2022seismic}& 0.9413 & 0.9233 & NGA & NGA & 0.9097 & NGA & 0.9327 & 0.9145 &  0.9361\\
Bilateral filter \cite{li2024spatial} & 0.9508 & 0.9413 & NGA & NGA & 0.9201 & NGA & 0.9406 & 0.9319 & 0.9486 \\
ANN \cite{novellino2021slow} & 0.8428 & 0.8058 & NGA & NGA & 0.7897 & NGA & -- & -- & -- \\
GBM \cite{novellino2021slow} & 0.8819 & 0.8501 & NGA & NGA & 0.7454 & NGA & -- & -- & -- \\
Ensemble \cite{rao2023earthquake} & -- & -- & -- & -- & -- & -- & 0.8619 & 0.8703 & 0.8555 \\
\bottomrule
\end{tabular}
\label{ROCrealCombined}
\end{center}
\end{table*}

The cross-event evaluation further substantiates the robustness of Spatial-VCBN. For the 2023 Turkey-Syria earthquake (Figure \ref{ROC}(d)), Spatial-VCBN achieves an impressive AUC of 0.9874 for building damage, significantly outperforming the prior (0.9051) with improvement of 9.1\%. Tables \ref{ROCrealCombined} and \ref{F1realCombined} reveal that this superiority is maintained across metrics, with our F1 score of 0.9412 substantially outperforming all alternatives including the Bilateral filter (0.9089). Notably, Spatial-VCBN demonstrates particularly strong improvements over learning-based methods such as the Ensemble approach, which achieves only F1 score of 0.8213 for the Turkey-Syria earthquake, highlighting the advantage of our causal framework over pure statistical learning approaches. Similar improvements are observed for the 2021 Haiti earthquake, with Spatial-VCBN achieving AUCs of 0.9703 for landslide (Figure \ref{ROC}(e)) and 0.9774 for building damage (Figure \ref{ROC}(f)), representing improvements of 26.0\% over the prior (0.7755) for building damage, and 5.15\% over the prior (0.9228) for landslide. Table \ref{ROCrealCombined} shows that even compared to the stronger Bilateral filter approach, Spatial-VCBN maintains advantages of 2.0\% and 3.9\% for landslide and building damage estimation, respectively. 

A comprehensive examination of Tables \ref{ROCrealCombined} and \ref{F1realCombined} reveals several additional insights not immediately apparent from the ROC curves alone. First, the performance gap between Spatial-VCBN and alternatives is consistently larger for building damage than for landslide estimation across all earthquake events, suggesting that our spatial-causal modeling approach particularly excels at capturing the complex mechanisms underlying building vulnerability. Second, machine learning methods like ANN and GBM show substantially worse performance than causal modeling approaches across all metrics, with AUC differences of up to 0.1475 (comparing Spatial-VCBN to ANN for landslide estimation in Puerto Rico) and F1 score differences of up to 0.2041 (comparing Spatial-VCBN to GBM for liquefaction estimation in Puerto Rico). This highlights the limitations of purely data-driven approaches when physical causal mechanisms are not explicitly modeled.

\begin{table*}[t]
\begin{center}
\caption{\textbf{F1 score comparison across earthquake events.} LS: Landslide, LF: Liquefaction, BD: Building Damage. HT: Haiti, PR: Puerto Rico, TK: Turkey. NGA: No Ground Truth Available. Bold values represent the best performance for each column.}
\begin{tabular}{l|ccc|ccc|ccc}
\toprule
\multirow{2}{*}{\textbf{Model}} & \multicolumn{3}{c|}{\textbf{Landslides}} & \multicolumn{3}{c|}{\textbf{Liquefaction}} & \multicolumn{3}{c}{\textbf{Building Damage}} \\
& HT & PR & TK & HT & PR & TK & HT & PR & TK \\
\midrule
\textbf{Spatial-VCBN} & \textbf{0.9297} & \textbf{0.9173} & NGA & NGA & \textbf{0.9107} & NGA & \textbf{0.9325} & \textbf{0.9167} & \textbf{0.9412} \\
Prior Model & 0.8845 & 0.8797 & NGA & NGA & 0.8301 & NGA & 0.7423 & 0.6756 & 0.8678 \\
VBCI & 0.9015 & 0.8864 & NGA & NGA & 0.8714 & NGA & 0.8947 & 0.8778 & 0.8973 \\
Bilateral filter & 0.9102 & 0.9031 & NGA & NGA & 0.8815 & NGA & 0.9023 & 0.8945 & 0.9089 \\
ANN & 0.8102 & 0.7732 & NGA & NGA & 0.7582 & NGA & -- & -- & -- \\
GBM & 0.8465 & 0.8160 & NGA & NGA & 0.7149 & NGA & -- & -- & -- \\
Ensemble & -- & -- & -- & -- & -- & -- & 0.8274 & 0.8356 & 0.8213 \\
\bottomrule
\end{tabular}
\label{F1realCombined}
\end{center}
\end{table*}

These results highlight three key strengths of Spatial-VCBN. First, Spatial-VCBN demonstrates consistent performance advantages across different hazard types, with particularly notable improvements for building damage estimation where distinguishing between different damage mechanisms is especially challenging. Second, Spatial-VCBN shows strong transferability across different earthquake events in diverse geographical settings, maintaining its performance edge in the Puerto Rico, Haiti, and Turkey-Syria earthquakes despite their varying geological contexts and built environment characteristics. Third, Spatial-VCBN achieves robust performance improvements particularly at low false positive rates, which is crucial for practical deployment in disaster response scenarios where false alarms can waste limited resources and undermine confidence in automated hazard assessments.
The consistent outperformance across multiple events, hazard types, and evaluation metrics demonstrates that modeling spatially-varying causal relationships through combined Gaussian Processes and normalizing flows provides a powerful framework for disaster impact assessment that generalizes well across different scenarios. This performance advantage is from the ability of Spatial-VCBN to capture both the spatial correlation structure inherent in disaster impacts and the complex, non-Gaussian distributions of causal effects that characterize real-world hazard-damage relationships. By learning location-specific causal parameters rather than assuming uniform relationships across the study area, Spatial-VCBN can adapt to the unique geological, structural, and environmental factors that influence how different hazards manifest in different locations.

\subsection{Ablation Study and Model Robustness}

\subsubsection{Computational Efficiency Analysis}

Table \ref{table:time} provides insights into the computational requirements of our framework across three earthquake events with varying affected areas. We implement a pruning strategy that maintains efficiency by focusing computational resources on active nodes while eliminating inactive ones from the processing pipeline. For the Haiti earthquake with the largest affected area (15,970 $km^2$), Spatial-VCBN required 15,029 seconds (approximately 4.2 hours), while the smaller Puerto Rico earthquake area (1,305 $km^2$) was processed in just 1,198 seconds (approximately 20 minutes). This near-linear scaling (with a processing rate of approximately 0.94 seconds per $km^2$) indicates that our framework maintains computational efficiency as the geographical scope increases.

The consistency in processing time per unit area across different earthquake events suggests that the computational demands of Spatial-VCBN are primarily determined by the spatial extent rather than being affected by the complexity or specific characteristics of different regions. This predictable scaling behavior is particularly valuable for emergency response scenarios where estimation of required computational resources is crucial for timely deployment. Our implementation leverages GPU acceleration, equipped with an NVIDIA Tesla T4 GPU (15 GB VRAM) and 51 GB of RAM, enabling rapid parallel processing of spatial data. The results indicate that even for large-scale events like the Turkey-Syria earthquake (4,676 $km^2$), analysis can be completed within reasonable timeframes (approximately 1.3 hours), making our approach practical for operational use in disaster response.

\begin{table}[t]
\centering
\caption{This table shows the time cost of running our framework using the same batch size in three earthquake events using real-world data.}
\begin{tabular}{c|ccc}
\toprule
\textbf{Method}  & \textbf{Haiti EQ.} & \textbf{ Puerto Rico EQ.} & \textbf{Turkey-Syria EQ.} \\
\hline
Map size & 15,970 $km^2$ & 1,305 $km^2$ & 4,676 $km^2$ \\
Time(s) & 15,029 & 1,198 & 4,543 \\
\bottomrule
\end{tabular}%
\label{table:time}
\end{table}

\subsubsection{Hyperparameter Sensitivity}

Table \ref{ab:K} presents a systematic evaluation of the sensitivity of Spatial-VCBN to the flow number $K$ in the normalizing flow component, which directly affects the expressiveness of the spatially-varying causal parameter distributions. The results reveal several important patterns across different earthquake events and hazard types. For all hazard types and earthquake events, performance improves substantially as $K$ increases from 2 to 6, with the significant improvements observed in the early stages. For example, in the Haiti earthquake building damage estimation, AUC increases from 0.8915 with $K = 2$ to 0.9774 with $K = 6$, an improvement of 9.6\%. This pattern indicates that the spatial distribution of causal parameters in disaster scenarios exhibits complexity that cannot be adequately captured by simpler flow architectures.

Notably, we observe that $K = 6$ represents an optimal balance point across all hazards and events, achieving the highest AUC values for landslide estimation in Haiti (0.9703), building damage in Haiti (0.9774), landslide in Puerto Rico (0.9527), liquefaction in Puerto Rico (0.9489), building damage in Puerto Rico (0.9532), and building damage in Turkey-Syria (0.9874). Beyond $K = 6$, we observe a slight performance degradation, with AUC decreasing marginally for $K = 7$ and $K = 9$. This pattern suggests a form of overfitting when the flow becomes too expressive relative to the available constraints.

The impact of $K$ varies across different hazard types, with building damage showing the highest sensitivity to flow complexity. For the Puerto Rico earthquake, increasing $K$ from 2 to 6 improves building damage AUC by 0.0818, compared to improvements of 0.0514 for landslide and 0.0593 for liquefaction. This difference likely reflects the more complex causal mechanisms involved in building damage, which depends on both the primary hazards and structural characteristics. Cross-event comparison reveals consistent patterns in hyperparameter sensitivity, with all three earthquake events showing similar optimal flow complexity despite their geographical and geological differences. This consistency suggests that Spatial-VCBN architecture captures fundamental aspects of the spatially-varying causal relationships in disaster scenarios rather than simply fitting to dataset-specific patterns.

\begin{table}[htbp]
\begin{center}
\caption{Ablation study on flow number $K$ in normalizing flows. NGA means no ground truth available. Bold values represent the best performance.}
\begin{tabular}{p{1.5cm}|p{1cm}p{1cm}p{1cm}|p{1cm}p{1cm}p{1cm}|p{1cm}p{1cm}p{1cm}}
\toprule
&  & Haiti EQ. &  &  & Puerto Rico EQ.  &  & & Turkey-Syria EQ. & \\
\hline
$K$ & \textbf{$AUC_{LS}$} & \textbf{$AUC_{LF}$} & \textbf{$AUC_{BD}$} & \textbf{$AUC_{LS}$} & \textbf{$AUC_{LF}$} & \textbf{$AUC_{BD}$} & \textbf{$AUC_{LS}$} & \textbf{$AUC_{LF}$} & \textbf{$AUC_{BD}$}\\
\hline
$K = 2$ & 0.9239 & NGA & 0.8915 & 0.9013 & 0.8896 & 0.8714 & NGA & NGA & 0.9089 \\
$K = 3$	& 0.9425 & NGA & 0.9178 & 0.9187 & 0.9057 & 0.8982 & NGA & NGA & 0.9302 \\
$K = 4$ & 0.9568 & NGA & 0.9452 & 0.9311 & 0.9257 & 0.9214 & NGA & NGA & 0.9587 \\
$K = 5$ & 0.9638 & NGA & 0.9641 & 0.9478 & 0.9375 & 0.9422 & NGA & NGA & 0.9755 \\
\textbf{$K = 6$} & \textbf{0.9703} & NGA & \textbf{0.9774} & \textbf{0.9527} & \textbf{0.9489} & \textbf{0.9532} & NGA & NGA & \textbf{0.9874} \\
$K = 7$	& 0.9701 & NGA & 0.9772 & 0.9525 & 0.9486 & 0.9530 & NGA & NGA & 0.9871 \\
$K = 9$	& 0.9695 & NGA & 0.9767 & 0.9522 & 0.9481 & 0.9527 & NGA & NGA & 0.9869 \\
\toprule
\end{tabular}
\label{ab:K}
\end{center}
\end{table}

\section{Discussion}\label{sec_discussion}

The development of a spatially-aware causal Bayesian network with normalizing flows represents a significant advancement in multi-hazard disaster impact assessment. Our results demonstrate that this approach not only improves estimation accuracy across multiple hazards and impacts induced by different earthquake events, with AUC improvements of up to 35.2\% over prior probability baselines and 5.5\% over state-of-the-art VBCI methods. Additionally, it provides interpretable insights into the complex causal mechanisms underlying disaster impacts. In this discussion, we contextualize our findings within the broader disaster science literature and examine their implications for both theoretical understanding and practical applications.

Traditional approaches to hazard assessment often rely on either purely data-driven methods that lack causal interpretability or physical models that struggle to incorporate the complex spatial heterogeneity of real-world disaster contexts. Our framework bridges this gap by explicitly modeling spatially-varying causal relationships through a combination of Gaussian Processes and normalizing flows. This innovation allows us to capture both the spatial correlation structure inherent in disaster impacts and the non-Gaussian, potentially multimodal distributions of causal effects that characterize real-world hazard-damage relationships.

The spatial patterns observed in our causal parameters ($\gamma_{LS}$, $\gamma_{LF}$, $\lambda_{LS}$, $\lambda_{LF}$, and $\lambda_{BD}$) highlight the fundamental importance of accounting for spatial heterogeneity in disaster modeling. For instance, in the Puerto Rico earthquake (Figure \ref{PR_gamma_posteriors}), the clear differentiation between landslide-dominated hillsides and liquefaction-dominated lowlands demonstrates how terrain characteristics fundamentally alter causal mechanisms. This spatial variability, along with the distinct patterns of building damage influence, suggests that assuming spatial homogeneity in causal relationships, as many existing models do, may lead to substantial inaccuracies in hazard assessment. The optimal flow number of $K = 6$ identified in our ablation study (Table \ref{ab:K}) suggests that real-world causal relationships in disaster contexts exhibit complexity that cannot be adequately captured by simpler distributional assumptions, underscoring the importance of our methodological innovation.

Remote sensing plays a crucial role in disaster response, but the interpretation of complex signals like DPM remains challenging. The ability of Spatial-VCBN to decouple multiple contributing factors to DPM signals addresses a significant gap in existing approaches. As demonstrated in Figure \ref{PRLSLFBDlambda}, our framework can identify distinct causal pathways from different hazard types to observed DPM signals within the same geographic region, providing critical insights that would be lost in approaches that treat DPM as a uniform indicator of damage. The results demonstrating robust causal parameter extraction even in signal-constrained environments (Figures \ref{PRLSlambda} and \ref{PRBDlambda}) highlight the potential of causal modeling to enhance the utility of remote sensing data in challenging conditions. For example, in mountainous regions where snow cover and complex topography introduce noise into DPM measurements, Spatial-VCBN successfully extracts coherent landslide susceptibility patterns that align with terrain characteristics visible in satellite imagery. The consistent performance across different hazard types and earthquake events suggests that our approach captures fundamental physical relationships rather than only fitting to dataset-specific patterns. This is particularly notable given the diverse geological and built environment contexts represented in our study areas (Puerto Rico, Haiti, and Turkey-Syria), as evidenced by the comparable AUC values achieved across these events (Table \ref{ROCrealCombined}). The transferability of Spatial-VCBN suggests that the spatially-varying causal relationships we identify reflect genuine physical processes that generalize across different disaster scenarios.

The practical implications of our work extend beyond theoretical advancements to offer tangible benefits for disaster risk reduction and emergency response. Particularly noteworthy is the strong performance of Spatial-VCBN at low false positive rates, as shown in the ROC curves (Figure \ref{ROC}). In practical disaster response, false alarms can waste limited resources and undermine stakeholder confidence in automated assessment systems. The ability of our approach to maintain high detection rates while minimizing false positives addresses this crucial operational concern. The reasonable computational requirements demonstrated in our efficiency analysis (Table \ref{table:time}), with processing times of approximately 0.94 seconds per $km^2$ with GPU acceleration, suggest that our approach is viable for operational deployment, even for large-scale events. The near-linear scaling with geographical area provides predictable resource requirements for emergency management agencies planning post-disaster assessments. For example, our results indicate that even a large-scale event like the Haiti earthquake (15,970 $km^2$) can be analyzed in approximately 4.2 hours using modest computational resources. Beyond immediate response applications, our spatially-explicit causal framework offers valuable insights for long-term disaster risk reduction planning. The identification of areas where specific hazard mechanisms dominate could inform targeted mitigation strategies. For example, authorities might prioritize slope stabilization in regions with high $\gamma_{LS}$ values or focus on liquefaction-resistant foundation designs in areas with elevated $\gamma_{LF}$ parameters.

 The spatially-varying causal parameters identified by our framework could serve as valuable calibration or validation data for detailed physical simulations, potentially improving their spatial accuracy. Conversely, insights from physical models could inform prior distributions in our Bayesian framework, creating a virtuous cycle of model improvement. This integration of data-driven causal inference with physical understanding represents a promising direction for advancing multi-hazard assessment methodologies. The ability of Spatial-VCBN to identify meaningful patterns even in areas with weak or noisy DPM signals suggests that it could extend the utility of physics-based models to regions where observational data is limited or compromised by environmental factors. This is particularly relevant for global-scale hazard assessment, where data quality varies substantially across different regions.

While our approach complements rather than replaces physics-based hazard models, the identified spatially-varying causal parameters could serve as valuable calibration data for physical simulations. Conversely, insights from physical models could inform prior distributions in our Bayesian framework. Despite promising results, limitations remain. Future work should validate our approach across a broader range of disaster types and incorporate temporal dynamics of hazard evolution.

In conclusion, our spatially-aware causal Bayesian network advances disaster impact assessment through its ability to capture spatially heterogeneous causal relationships, maintain robust performance under varying signal conditions, and provide interpretable insights into complex disaster mechanisms. The framework addresses critical limitations in existing approaches that either lack causal interpretability \cite{novellino2021slow, rao2023earthquake} or struggle with spatial heterogeneity \cite{xu2022seismic, li2023disasternet}. Our integration of Gaussian Processes with normalizing flows builds upon advances in spatially-varying coefficient models \cite{banerjee2003hierarchical} while extending their applicability to non-Gaussian posterior distributions. The performance improvements demonstrated across multiple hazard types and earthquake events, particularly the substantial gains in building damage assessment, validate the practical utility of our approach for disaster response operations.

\section{Declaration Statements}
\subsection{Data availability}

Data used in this study were collected from several publicly accessible sources. The primary observational data consists of Damage Proxy Maps (DPMs) generated by NASA's Advanced Rapid Imaging and Analysis (ARIA) team using InSAR imagery from Sentinel-1 satellites, available at \url{https://aria-share.jpl.nasa.gov/}. These DPMs capture correlation changes between pre- and post-event images, providing valuable information for rapid hazard and impact estimation.

For ground truth validation, we collected data from multiple sources across the three earthquake events studied. For the 2021 Haiti earthquake (M7.2), building damage and landslide inventories were provided by StEER (available at: \url{https://www.steer.network/haiti-response}) and GEER teams. Field reconnaissance data for the 2020 Puerto Rico earthquake (M6.4) was collected by USGS, University of Puerto Rico Mayagüez, GEER, and StEER teams, available at \url{https://www.sciencebase.gov/catalog/item/5eb5b9dc82ce25b5135ae83a}. For the 2023 Turkey-Syria earthquake sequence (M7.8), we utilized building damage inventory data from the Turkish Ministry of Environment accessible at \cite{turkeyadm}.

Additional data sources used in this study include USGS ShakeMap and ground failure models (\url{https://earthquake.usgs.gov/}) as prior models, along with building footprints from OpenStreetMap (\url{https://www.openstreetmap.org/}). Spatial-VCBN evaluation utilized both synthesized data (generated from these real-world sources) and direct real-world observations. Any data not available through these public repositories may be obtained from the corresponding author upon reasonable request.

\subsection{Code Availability}
The underlying code and training/validation datasets for this study are available in the repository and can be accessed via \url{https://github.com/PaperSubmissionFinal/SpatialBN}.

\subsection{Acknowledgments}
The author(s) disclosed receipt of the following financial support for the research, authorship, and/ or publication of this article: X. L., S.G., R.G., and S. X. are supported by U.S. Geological Survey Grant G22AP00032 and NSF CMMI-2242590. 
Any mention of commercial products is for informational purposes and does not constitute an endorsement by the U.S. government.

\subsection{Author Contributions}
X.L., S.G., and S.X. conceptualized the research, developed the framework. R.G. conducted the data cleaning. X.L., S.G., R.G. implemented the code. X.L. and S.G. conducted the experiments, analyzed the results. S.X. acquired the funding to support this research. All authors wrote, read, and approved the final manuscript.

\subsection{Competing Interests}
The authors declare no competing interests.


\printbibliography 

@techreport{danielson2011global,
  title={Global multi-resolution terrain elevation data 2010 (GMTED2010)},
  author={Danielson, Jeffrey J and Gesch, Dean B},
  year={2011},
  institution={US Geological Survey}
}

@article{nowicki2018global,
  title={A global empirical model for near-real-time assessment of seismically induced landslides},
  author={Nowicki Jessee, MA and Hamburger, Michael W and Allstadt, Kate and Wald, David J and Robeson, Scott M and Tanyas, Hakan and Hearne, Mike and Thompson, Eric M},
  journal={Journal of Geophysical Research: Earth Surface},
  volume={123},
  number={8},
  pages={1835--1859},
  year={2018},
  publisher={Wiley Online Library}
}

@misc{arino2012global,
  title={Global land cover map for 2009 (GlobCover 2009)},
  author={Arino, Olivier and Ramos Perez, Jose Julio and Kalogirou, Vasileios and Bontemps, Sophie and Defourny, Pierre and Van Bogaert, Eric},
  year={2012},
  publisher={European Space Agency (ESA) \& Universit{\'e} catholique de Louvain (UCL~…}
}

@article{hartmann2012new,
  title={The new global lithological map database GLiM: A representation of rock properties at the Earth surface},
  author={Hartmann, Jens and Moosdorf, Nils},
  journal={Geochemistry, Geophysics, Geosystems},
  volume={13},
  number={12},
  year={2012},
  publisher={Wiley Online Library}
}

@article{ghofrani2014site,
  title={Site condition evaluation using horizontal-to-vertical response spectral ratios of earthquakes in the NGA-West 2 and Japanese databases},
  author={Ghofrani, Hadi and Atkinson, Gail M},
  journal={Soil Dynamics and Earthquake Engineering},
  volume={67},
  pages={30--43},
  year={2014},
  publisher={Elsevier}
}

@article{toprak2003liquefaction,
  title={Liquefaction potential index: field assessment},
  author={Toprak, Selcuk and Holzer, Thomas L},
  journal={Journal of Geotechnical and Geoenvironmental Engineering},
  volume={129},
  number={4},
  pages={315--322},
  year={2003},
  publisher={American Society of Civil Engineers}
}

@article{marc2016seismologically,
  title={A seismologically consistent expression for the total area and volume of earthquake-triggered landsliding},
  author={Marc, Odin and Hovius, Niels and Meunier, Patrick and Gorum, Tolga and Uchida, Taro},
  journal={Journal of Geophysical Research: Earth Surface},
  volume={121},
  number={4},
  pages={640--663},
  year={2016},
  publisher={Wiley Online Library}
}

@article{newmark1965effects,
  title={Effects of earthquakes on dams and embankments},
  author={Newmark, Nathan Mortimore},
  journal={Geotechnique},
  volume={15},
  number={2},
  pages={139--160},
  year={1965},
  publisher={Thomas Telford Ltd}
}

@article{zhu2016updated,
  title={Updated geospatial liquefaction model for global use},
  author={Zhu, Jing and Baise, LG and Thompson, EM},
  journal={Seismol. Res. Lett},
  volume={87},
  year={2016}
}

@article{beven1979physically,
  title={A physically based, variable contributing area model of basin hydrology/Un mod{\`e}le {\`a} base physique de zone d'appel variable de l'hydrologie du bassin versant},
  author={Beven, Keith J and Kirkby, Michael J},
  journal={Hydrological sciences journal},
  volume={24},
  number={1},
  pages={43--69},
  year={1979},
  publisher={Taylor \& Francis}
}

@article{yun2015rapid,
  title={Rapid damage mapping for the 2015 M w 7.8 Gorkha earthquake using synthetic aperture radar data from COSMO--SkyMed and ALOS-2 Satellites},
  author={Yun, Sang-Ho and Hudnut, Kenneth and Owen, Susan and Webb, Frank and Simons, Mark and Sacco, Patrizia and Gurrola, Eric and Manipon, Gerald and Liang, Cunren and Fielding, Eric and others},
  journal={Seismological Research Letters},
  volume={86},
  number={6},
  pages={1549--1556},
  year={2015},
  publisher={Seismological Society of America}
}

@article{kongar2017evaluating,
  title={Evaluating simplified methods for liquefaction assessment for loss estimation},
  author={Kongar, Indranil and Rossetto, Tiziana and Giovinazzi, Sonia},
  journal={Natural Hazards and Earth System Sciences},
  volume={17},
  number={5},
  pages={781--800},
  year={2017},
  publisher={Copernicus GmbH}
}

@article{Xu2012ComparisonOD,
  title={Comparison of different models for susceptibility mapping of earthquake triggered landslides related with the 2008 Wenchuan earthquake in China},
  author={Chong Xu and Xiwei Xu and Fuchu Dai and Arun K. Saraf},
  journal={Comput. Geosci.},
  year={2012},
  volume={46},
  pages={317-329},
  url={https://api.semanticscholar.org/CorpusID:154627}
}

@misc{Fig3,
  author = {{U.S. Geological Survey}},
  title = {M 7.8 - 26 km ENE of Nurdagi, Turkey - Ground Failure},
  howpublished = {Online},
url ={https://earthquake.usgs.gov/earthquakes/eventpage/us6000jllz/ground-failure/summary.},
  year = {2023}
}

@misc{ARIA1,
  title = {Advanced Rapid Imaging and Analysis (ARIA) - Center for Natural Hazards.},
  author = {{ARIA Data Share}},
  howpublished = {Online},
  url ={https://aria-share.jpl.nasa.gov/.},
  year = {2020}
}

@article{fawcett2006introduction,
  title={An introduction to ROC analysis},
  author={Fawcett, Tom},
  journal={Pattern recognition letters},
  volume={27},
  number={8},
  pages={861--874},
  year={2006},
  publisher={Elsevier}
}

@misc{turkeyadm,
  doi = {10.17603/DS2-7RY2-GV66},
  url = {https://www.designsafe-ci.org/data/browser/public/designsafe.storage.published/PRJ-3824v2/#details-942732811040452115-242ac11b-0001-012},
  author = {Dilsiz, Abdullah and Gunay, Selim and Mosalam, Khalid and Miranda, Eduardo and Arteta, Carlos and Sezen, Halil and Fischer, Erica and Hakhamaneshi, Manouchehr and Hassan, Wael and ALhawamdeh, Bilal and Andrus, Samuel and Archbold, Jorge and Arslanturkoglu, Safak and BEKTAS, NURULLAH and Ceferino, Luis and Cohen, Jade and Duran, Burak and Erazo, Kalil and Faraone, Gloria and Feinstein, Tali and Gautam, Rajendra and Gupta, Abhineet and Haj Ismail, Salah and Jana, Amalesh and Javadinasab Hormozabad, Sajad and Kasalanati, Amarnath and Kenawy, Maha and Khalil, Zeyad and Liou, Irene and Marinkovic, Marko and Martin, Amory and Merino-Peña, Yvonne and Mivehchi, Maziar and Moya, Luis and Pájaro Miranda, César and quintero, nicolas and Rivera, Juliana and Romão, Xavier and Lopez Ruiz, Maria Camila and Sorosh, Shokrullah and Vargas, Laura and Velani, Pulkit Dilip and Wibowo, Hartanto and Xu, Susu and YILMAZ, TANER and Alam, Mohammad and Holtzer, Gabor and Kijewski-Correa, Tracy and Robertson, Ian and Roueche, David and Safiey, Amir},
  language = {en},
  title = {StEER: 2023 Mw 7.8 Kahramanmaras, Türkiye Earthquake Sequence Preliminary Virtual Reconnaissance Report (PVRR)},
  publisher = {Designsafe-CI},
  year = {2023}
}

@inproceedings{Banerjee2003HierarchicalMA,
  title={Hierarchical Modeling and Analysis for Spatial Data},
  author={Sudipto Banerjee and Bradley P. Carlin and Alan E. Gelfand},
  year={2003},
  url={https://api.semanticscholar.org/CorpusID:62708858}
}

@inproceedings{Diggle2013ModelbasedG,
  title={Model-based geostatistic},
  author={Peter John Diggle and Paulo Justiniano Ribeiro},
  year={2013},
  url={https://api.semanticscholar.org/CorpusID:134130701}
}

@article{Tomasi1998BilateralFF,
  title={Bilateral filtering for gray and color images},
  author={Carlo Tomasi and Roberto Manduchi},
  journal={Sixth International Conference on Computer Vision (IEEE Cat. No.98CH36271)},
  year={1998},
  pages={839-846},
  url={https://api.semanticscholar.org/CorpusID:14308539}
}

@inproceedings{Paris2009BilateralFT,
  title={Bilateral Filtering: Theory and Applications: Series: Foundations and Trends in Computer Graphics and Vision},
  author={Sylvain Paris and Pierre Kornprobst and Jack Tumblin and Fr{\'{e}}do Durand and Brian Curless and Luc Van Gool and Richard Szeliski},
  year={2009},
  url={https://api.semanticscholar.org/CorpusID:67417024}
}

@article{Gelfand2016SpatialSA,
  title={Spatial statistics and Gaussian processes: A beautiful marriage},
  author={Alan E. Gelfand and Erin M. Schliep},
  journal={spatial statistics},
  year={2016},
  volume={18},
  pages={86-104},
  url={https://api.semanticscholar.org/CorpusID:62363233}
}

@inproceedings{jang2009rescue,
  title={Rescue information system for earthquake disasters based on MANET emergency communication platform},
  author={Jang, Hung-Chin and Lien, Yao-Nan and Tsai, Tzu-Chieh},
  booktitle={Proceedings of the 2009 international conference on wireless communications and mobile computing: connecting the world wirelessly},
  pages={623--627},
  year={2009}
}

@inproceedings{havenith2022first,
  title={First analysis of landslides triggered by the August 14, 2021, Nippes (Haiti) earthquake, compared with the 2010 event},
  author={Havenith, Hans-Balder and Guerrier, Kelly and Schl{\"o}gel, Romy and Mreyen, Anne-Sophie and Ulysse, Sophia and Braun, Anika and Victor, Karl-Henry and Saint-Fleur, Newdeskarl and Cauchie, Lena and Boisson, Dominique and others},
  booktitle={EGU General Assembly Conference Abstracts},
  pages={EGU22--1980},
  year={2022}
}

@misc{li2023m7,
  title={M7. 8 Turkey-Syria Earthquake Impact Estimates from Near-real-time Crowdsourced and Remote Sensing Data},
  author={Li, Xuechun and Dimasaka, Joshua and Zhang, Xiaojian and Yu, Xiao and Wang, Chenguang and Noh, Hae Young and Hu, Xie and Zhao, Xilei and Xu, Susu},
  year={2023},
  publisher={DesignSafe-CI}
}

@article{wang2024scalable,
  title={Scalable and rapid building damage detection after hurricane Ian using causal Bayesian networks and InSAR imagery},
  author={Wang, Chenguang and Liu, Yepeng and Zhang, Xiaojian and Li, Xuechun and Paramygin, Vladimir and Sheng, Peter and Zhao, Xilei and Xu, Susu},
  journal={International Journal of Disaster Risk Reduction},
  volume={104},
  pages={104371},
  year={2024},
  publisher={Elsevier}
}

@inproceedings{wang2023causality,
  title={Causality-informed Rapid Post-hurricane Building Damage Detection in Large Scale from InSAR Imagery},
  author={Wang, Chenguang and Liu, Yepeng and Zhang, Xiaojian and Li, Xuechun and Paramygin, Vladimir and Subgranon, Arthriya and Sheng, Peter and Zhao, Xilei and Xu, Susu},
  booktitle={Proceedings of the 8th ACM SIGSPATIAL International Workshop on Security Response using GIS},
  pages={7--12},
  year={2023}
}

@inproceedings{xucausality,
  title={Causality-informed Bayesian inference for rapid seismic ground failure and building damage estimation},
  author={Wald, David J and Xu, Susu and Dimasaka, J and Noh, H},
  booktitle={12th National Conference on Earthquake Engineering},
  year={2023}
}

@article{miranda2020,
  title={StEER-07 Jan. 2020 Puerto Rico mw6. 4 Earthquake: preliminary virtual reconnaissance report (PVRR)},
  author={Miranda, Eduardo and Acosta, Andres and Aponte-Bermudez, Luis David and Roueche, David},
  journal={DesignSafe-CI. https://doi. org/10.17603/ds2-xfhz-fz88},
  year={2020}
}

@article{miranda2020b,
  title={StEER-Puerto Rico Earthquake Sequence December 2019 to January 2020: early access reconnaissance report (EARR)},
  author={Miranda, E and Archbold, J and Heresi, P and Messina, A and Rosa, I and Kijewski-Correa, T and Mosalam, K and Prevatt, D and Robertson, I and Roueche, D},
  journal={Designsafe-CI},
  year={2020}
}

@online{HaitiGEER,
  title = {2021 HAITI EARTHQUAKE},
  url ={https://geerassociation.org/component/geer_reports/?view=geerreports&id=100&layout=build.},
  author = {Geotechnical Extreme Events Reconnaissance},
  organization = {GEER},
  year = {2021}
}

@online{HaitiBDGT,
  title = {StEER's Preliminary Virtual Reconnaissance Report for M7.2 Nippes, Haiti earthquake released.},
  url ={https://www.steer.network/haiti-response.},
  author = {StEER Network},
  organization = {NSF}
}

@article{zhao2022evaluation,
  title={Evaluation of factors controlling the spatial and size distributions of landslides, 2021 Nippes earthquake, Haiti},
  author={Zhao, Bo and Wang, Yunsheng and Li, Weile and Lu, Huiyan and Li, Zhengyou},
  journal={Geomorphology},
  volume={415},
  pages={108419},
  year={2022},
  publisher={Elsevier}
}

@article{rao2023earthquake,
  title={Earthquake building damage detection based on synthetic-aperture-radar imagery and machine learning},
  author={Rao, Anirudh and Jung, Jungkyo and Silva, Vitor and Molinario, Giuseppe and Yun, Sang-Ho},
  journal={Natural Hazards and Earth System Sciences},
  volume={23},
  number={2},
  pages={789--807},
  year={2023},
  publisher={Copernicus GmbH}
}

@article{novellino2021slow,
  title={Slow-moving landslide risk assessment combining Machine Learning and InSAR techniques},
  author={Novellino, A and Cesarano, M and Cappelletti, P and Di Martire, D and Di Napoli, M and Ramondini, M and Sowter, A and Calcaterra, D},
  journal={Catena},
  volume={203},
  pages={105317},
  year={2021},
  publisher={Elsevier}
}

@techreport{fema2003hazus,
  title={HAZUS-MH Technical Manual},
  author={{Federal Emergency Management Agency}},
  year={2003},
  institution={Department of Homeland Security Emergency Preparedness and Response Directorate},
  address={Washington, D.C.},
  number={FEMA-433},
  type={Technical Manual}
}

@article{xu2022seismic,
  title={Seismic multi-hazard and impact estimation via causal inference from satellite imagery},
  author={Xu, Susu and Dimasaka, Joshua and Wald, David J and Noh, Hae Young},
  journal={Nature Communications},
  volume={13},
  number={1},
  pages={7793},
  year={2022},
  publisher={Nature Publishing Group UK London}
}

@article{li2024spatial,
  title={Spatial-variant causal Bayesian inference for rapid seismic ground failures and impacts estimation},
  author={Li, Xuechun and Xu, Susu},
  journal={arXiv preprint arXiv:2412.00026},
  year={2024}
}

@article{ray2019bayesian,
  title={Bayesian geophysical inversion with trans-dimensional Gaussian process machine learning},
  author={Ray, Anandaroop and Myer, David},
  journal={Geophysical Journal International},
  volume={217},
  number={3},
  pages={1706--1726},
  year={2019},
  publisher={Oxford University Press}
}

@article{li2025rapid,
  title={Rapid building damage estimates from the M7. 8 Turkey earthquake sequence via causality-informed Bayesian inference from Satellite Imagery},
  author={Li, Xuechun and Yu, Xiao and B{\"u}rgi, Paula M and Wald, David J and Hu, Xie and Xu, Susu},
  journal={Earthquake Spectra},
  volume={41},
  number={1},
  pages={5--33},
  year={2025},
  publisher={SAGE Publications Sage UK: London, England}
}

@inproceedings{li2023disasternet,
  title={Disasternet: Causal Bayesian networks with normalizing flows for cascading hazards estimation from satellite imagery},
  author={Li, Xuechun and B{\"u}rgi, Paula M and Ma, Wei and Noh, Hae Young and Wald, David Jay and Xu, Susu},
  booktitle={Proceedings of the 29th ACM SIGKDD Conference on Knowledge Discovery and Data Mining},
  pages={4391--4403},
  year={2023}
}

@inproceedings{rezende2015variational,
  title={Variational inference with normalizing flows},
  author={Rezende, Danilo and Mohamed, Shakir},
  booktitle={International conference on machine learning},
  pages={1530--1538},
  year={2015},
  organization={PMLR}
}

@article{yu2024intelligent,
  title={Intelligent assessment of building damage of 2023 Turkey-Syria Earthquake by multiple remote sensing approaches},
  author={Yu, Xiao and Hu, Xie and Song, Yuqi and Xu, Susu and Li, Xuechun and Song, Xiaodong and Fan, Xuanmei and Wang, Fang},
  journal={NPJ Natural Hazards},
  volume={1},
  number={1},
  pages={3},
  year={2024},
  publisher={Nature Publishing Group UK London}
}

@inproceedings{li2024optimizing,
  title={Optimizing Rapid Seismic Building Damage Assessment: Integrating Enhanced Radar Change Detection Maps with Variational Bayesian Networks},
  author={Li, Xuechun and Gao, Runyu and Burgi, Paula M and Wald, David J and Xu, Susu},
  booktitle={IGARSS 2024-2024 IEEE International Geoscience and Remote Sensing Symposium},
  pages={3791--3795},
  year={2024},
  organization={IEEE}
}

@article{li2023normalizing,
  title={Normalizing flow-based deep variational bayesian network for seismic multi-hazards and impacts estimation from insar imagery},
  author={Li, Xuechun and Burgi, Paula M and Ma, Wei and Noh, Hae Young and Wald, David J and Xu, Susu},
  journal={arXiv preprint arXiv:2310.13805},
  year={2023}
}

@article{farr2007shuttle,
  title={The shuttle radar topography mission},
  author={Farr, Tom G and Rosen, Paul A and Caro, Edward and Crippen, Robert and Duren, Riley and Hensley, Scott and Kobrick, Michael and Paller, Mimi and Rodriguez, Ernesto and Roth, Ladislav and others},
  journal={Reviews of geophysics},
  volume={45},
  number={2},
  year={2007},
  publisher={Wiley Online Library}
}

@article{wald1999relationships,
  title={Relationships between peak ground acceleration, peak ground velocity, and modified Mercalli intensity in California},
  author={Wald, David J and Quitoriano, Vincent and Heaton, Thomas H and Kanamori, Hiroo},
  journal={Earthquake spectra},
  volume={15},
  number={3},
  pages={557--564},
  year={1999},
  publisher={SAGE Publications Sage UK: London, England}
}

@article{seyhan2014semi,
  title={Semi-empirical nonlinear site amplification from NGA-West2 data and simulations},
  author={Seyhan, Emel and Stewart, Jonathan P},
  journal={Earthquake Spectra},
  volume={30},
  number={3},
  pages={1241--1256},
  year={2014},
  publisher={SAGE Publications Sage UK: London, England}
}

@article{kamp2008gis,
  title={GIS-based landslide susceptibility mapping for the 2005 Kashmir earthquake region},
  author={Kamp, Ulrich and Growley, Benjamin J and Khattak, Ghazanfar A and Owen, Lewis A},
  journal={Geomorphology},
  volume={101},
  number={4},
  pages={631--642},
  year={2008},
  publisher={Elsevier}
}

@article{lehner2006hydrosheds,
  title={HydroSHEDS technical documentation},
  author={Lehner, Bernhard and Verdin, Kris and Jarvis, Andy},
  journal={World Wildlife Fund US, Washington, DC},
  volume={5},
  year={2006}
}

@article{wald2003shakemap,
  title={ShakeMap, a tool for earthquake response},
  author={Wald, David and Wald, Lisa and Worden, Bruce and Goltz, Jim},
  journal={Fact Sheet},
  year={2003},
  publisher={US Geological Survey}
}

@book{banerjee2003hierarchical,
  title={Hierarchical modeling and analysis for spatial data},
  author={Banerjee, Sudipto and Carlin, Bradley P and Gelfand, Alan E},
  year={2003},
  publisher={Chapman and Hall/CRC}
}

@misc{aria2020pr,
    author = {{Advanced Rapid Imaging and Analysis (ARIA) - Center for Natural Hazards}},
    title = {{ARIA Data Share}},
    url = {https://aria-share.jpl.nasa.gov/20200106-Puerto\_Rico\_EQ/DPM/} ,
    year = 2020
}

@online{priorHaiti,
  title = {United States Geological Survey. M 7.2 - Nippes, Haiti.},
  author = {U.S. Geological Survey.},
  url = {https://earthquake.usgs.gov/earthquakes/eventpage/us6000f65h/executive.},
  year = {2021},
  organization = {USGS}
}

@online{reportHaiti,
  title = {Haiti: Earthquake Situation Report No. 4 (7 September 2021).},
  author = {OCHA Haiti.},
  url = {https://digitalcommons.fiu.edu/cgi/viewcontent.cgi?article=1854&context=srhreports},
  year = {2021},
  organization = {OCHA}
}

@article{allstadt2022ground,
  author    = {Allstadt, K.E. and Thompson, E.M. and García, D.B. and Brugman, E.I. and Hughes, K.S. and Schmitt, R.G.},
  title     = {Ground Failure Triggered by the 7 January 2020 Mw 6.4 Puerto Rico Earthquake},
  journal   = {Seismological Research Letters},
  year      = {2022}
}

@article{fielding2024damage,
  title={Damage Proxy Mapping with SAR interferometric coherence change},
  author={Fielding, Eric Jameson and Jung, Jungkyo},
  journal={Procedia Computer Science},
  volume={239},
  pages={2322--2328},
  year={2024},
  publisher={Elsevier}
}





\end{document}